\documentclass[runningheads,table]{llncs}
\usepackage[table]{xcolor}
\usepackage{graphicx}

\usepackage{tikz}
\usepackage{comment}
\usepackage{amsmath,amssymb} %
\usepackage{booktabs}
\usepackage{pifont}
\usepackage{makecell}
\usepackage{caption}
\usepackage{subcaption}
\usepackage{stfloats}
\usepackage{wrapfig}
\usepackage{adjustbox}
\usepackage[para]{footmisc}
\usepackage{multirow}
\usepackage{enumitem}
\usepackage{wrapfig}
\usepackage[pagebackref,breaklinks,colorlinks]{hyperref}
\usepackage[capitalize]{cleveref}
\usepackage{cite}
\crefname{section}{Sec.}{Secs.}
\Crefname{section}{Section}{Sections}
\Crefname{table}{Table}{Tables}
\crefname{table}{Tab.}{Tabs.}
\newcommand{\cmark}{\color{green}{\ding{51}}}%
\newcommand{\xmark}{\color{red}{\ding{55}}}%
\newcommand{\etal}{\textit{et al}.} 
\newcommand{\ie}{\textit{i}.\textit{e}. }

\usepackage[font=small,labelfont=bf]{caption}
\usepackage[accsupp]{axessibility}  %
\usepackage{breakurl}

\definecolor{darkgreen}{RGB}{30,150,30}

\begin{document}

\pagestyle{headings}
\mainmatter
\def\ECCVSubNumber{2334}  %

\title{Improving GANs for Long-Tailed Data through Group Spectral Regularization} %

\titlerunning{gSR: group Spectral Regularizer for GANs}
\author{Harsh Rangwani\inst{1} \and
Naman Jaswani\inst{1} \and
Tejan Karmali\inst{1,2} \and \\
Varun Jampani\inst{2} \and
R. Venkatesh Babu\inst{1}
}
\authorrunning{Rangwani et al.}
\institute{Indian Institute of Science, Bengaluru, India 
\\ \and
Google Research}

\maketitle

\begin{abstract}
Deep long-tailed learning aims to train useful deep networks on practical, real-world imbalanced distributions, wherein most labels of the tail classes are associated with a few samples. There has been a large body of work to train discriminative models for visual recognition on long-tailed distribution. In contrast, we aim to train conditional Generative Adversarial Networks, a class of image generation models on long-tailed distributions. We find that similar to recognition, state-of-the-art methods for image generation also suffer from performance degradation on tail classes. The performance degradation is mainly due to class-specific mode collapse for tail classes, which we observe to be correlated with the spectral explosion of the conditioning parameter matrix. We propose a novel group Spectral Regularizer (gSR) that prevents the spectral explosion alleviating mode collapse, which results in diverse and plausible image generation even for tail classes. We find that gSR effectively combines with existing augmentation and regularization techniques, leading to state-of-the-art image generation performance on long-tailed data. Extensive experiments demonstrate the efficacy of our regularizer on long-tailed datasets with \mbox{different} degrees of imbalance. Project Page: \href{https://sites.google.com/view/gsr-eccv22}{https://sites.google.com/view/gsr-eccv22}.  

\end{abstract}

\section{Introduction}
\label{sec:intro}
Generative Adversarial Networks (GAN)~\cite{goodfellow2014generative} are consistently at the forefront of generative models for image distributions, also being used for diverse applications like image-to-image translation~\cite{mao2019mode}, super resolution~\cite{ledig2017photo} etc. One of the classical applications of GAN is the class specific image generation by conditioning on the class label $y$. The generated images in ideal case should associate to class label $y$, be of high quality and exhibit diversity. The conditioning is usually achieved with conditional Batch Normalization (cBN)~\cite{de2017modulating} layers which induce class-specific ($y$) features at each layer of the generator. The additional class conditioning information enables GAN models like the state-of-the-art (SOTA) BigGAN~\cite{brock2018large} to generate diverse images, in comparison to unconditional models~\cite{kang2020contrastive}.

Recent works~\cite{zhao2020differentiable} demonstrate
that performance of models like BigGAN deteriorates from mode collapse when limited training data is presented. The differentiable data augmentation \mbox{approaches}~\cite{zhao2020differentiable, karras2020training, tran2021data} attempt to mitigate this degradation by enriching the training data through augmentations. On the other hand, model based regularization techniques like LeCam~\cite{tseng2021regularizing} are also proposed to prevent the degradation of image quality in such cases.

\begin{figure*}[t]
    \centering
    
    \includegraphics[width=\textwidth,height=4.5cm]{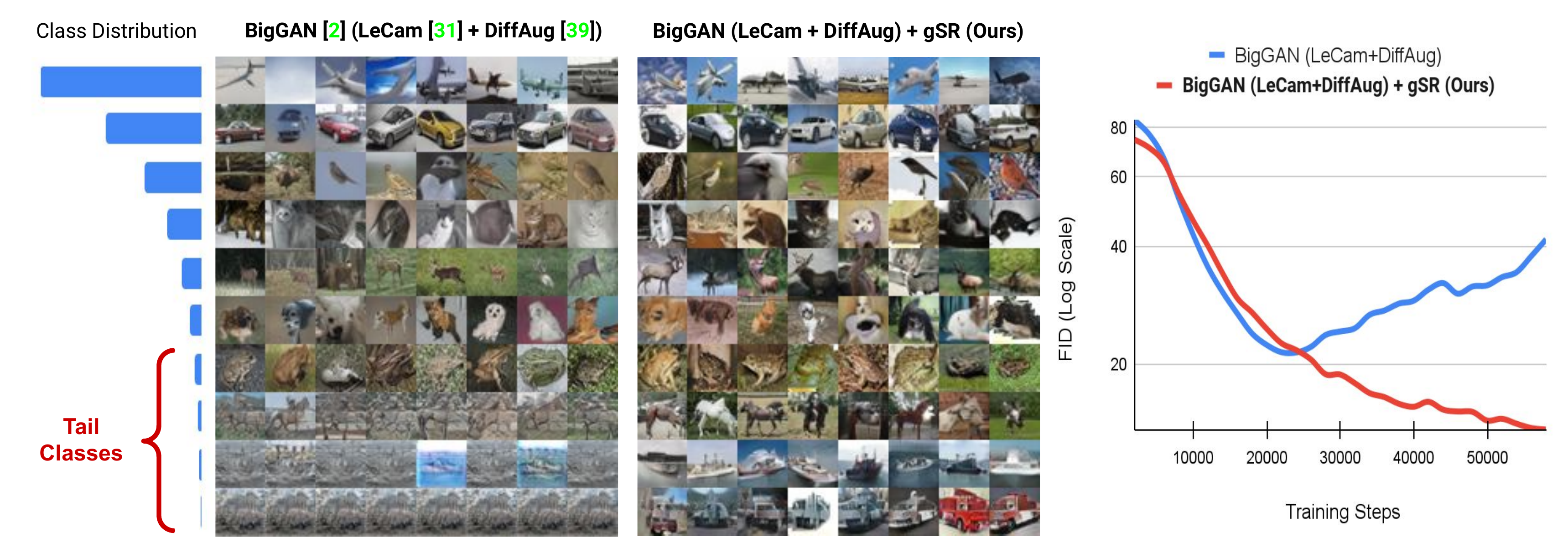}
    \caption{\textbf{Regularizing GANs on long-tailed training data.} \emph{(left)} Images generated from BigGAN trained on long-tailed CIFAR-10. \emph{(right)} FID scores vs. training steps. The proposed gSR regularizer prevents mode collapse, for the tail classes~\cite{brock2018large, tseng2021regularizing, zhao2020differentiable}. }\label{fig:overview}
\end{figure*}%
Although these methods lead to effective improvement in image generation quality, they are designed to increase performance when trained on balanced datasets (\ie even distribution of samples across classes). We find that the SOTA methods like BigGAN  with LeCam and augmentation, which are designed for limited data, also suffer from the phenomenon of class-specific mode collapse when trained on long-tailed datasets. By \textit{class-specific mode collapse}, we refer to deteriorating quality of generated images for tail classes, as shown in Fig.~\ref{fig:overview}.

In this work we aim to investigate the cause of the class-specific mode collapse that is observed in the generated images of tail classes. We find that the class-specific mode collapse is correlated with spectral explosion (\ie sudden increase in spectral norms, ref. Fig.~\ref{fig:sn_fid}) of the corrosponding class-specific (cBN) parameters (when grouped into a matrix, described in Sec.~\ref{sec:regularizer}). %
To prevent this spectral explosion, we introduce a novel class-specific \textit{group \mbox{Spectral Regularizer}} (gSR), which constrains the spectral norm of class-specific cBN parameters. Although there are multiple spectral~\cite{yoshida2017spectral, vahdat2020NVAE} regularization (and normalization~\cite{miyato2018spectral}) techniques used in deep learning literature, all of them are specific to model weights $W$, whereas our regularizer focuses on cBN parameters. We, through our analysis, show that our proposed gSR leads to reduced correlation among tail classes' cBN parameters, effectively mitigating the issue of class-specific mode collapse. 

We extensively test our regularizer's effectiveness by combining it with popular SNGAN~\cite{miyato2018spectral} and BigGAN~\cite{brock2018large} architectures. It also complements discriminator specific SOTA regularizer (LeCam+DiffAug~\cite{tseng2021regularizing}), as it's combination with gSR ensures improved quality of image generation even across tail classes (Fig.~\ref{fig:overview}). In summary, we make the following contributions:
\begin{itemize}
\itemsep0em 
    \item We first report the phenomenon of class-specific mode collapse, which is observed when cGANs are trained on long-tailed imbalance datasets. We find that spectral norm explosion of class-specific cBN parameters correlates with its mode collapse.
    \item We find that even existing techniques for limited data~\cite{tseng2021regularizing, liu2021generative, Karras2020ada} are unable to prevent class-specific collapse. Hence, we propose a novel group Spectral Regularizer (gSR) which helps in alleviating such collapse.
    \item   Combining gSR with existing SOTA GANs leads to large average relative improvement (of $\sim25\%$ in FID) for image generation on 5 different long-tailed dataset configurations.
\end{itemize}

\section{Related Work}
\noindent\textbf{Generative Adversarial Networks:} 
Generative Adversarial Networks~\cite{goodfellow2014generative} are a combination of Generator $G$ and Discriminator $D$ aimed at learning a generative model for a distribution. GANs have enabled learning of models for complex distributions like high resolution images etc. One of the inflection point for success was the invention of Spectral Normalization (SN) for GANs (SNGAN) which enabled GANs to scale to datasets like ImageNet~\cite{deng2009imagenet} (1000 classes). The GAN training was further scaled by BigGAN \cite{brock2018large} which demonstrated successful high resolution image generation, using a deep ResNet network. 

\noindent \textbf{Regularization:} Several regularization techniques \cite{zhang2019consistency, zhao2021improved, mao2019mode, kavalerov2019cgans, liu2019spectral, zhou2021omni} are developed to alleviate the problem of mode collapse in GANs like Gradient Penalty~\cite{gulrajani2017improved}, Spectral Normalization~\cite{miyato2018spectral} etc. These include LeCam~\cite{tseng2021regularizing} and Differentiable Augmentations~\cite{zhao2020differentiable, Karras2020ada} which are the regularization techniques specifically designed to prevent mode collapse in limited data scenarios. Commonality among majority of these techniques are that they (a) designed for the data which is balanced across classes, and (b) focus on discriminator network $D$. In this work, we aim to regularize the cBN parameters in generator $G$, which makes our regularizer complementary to earlier works.

\noindent \textbf{Long-Tailed Imbalanced Data:} Long-tailed imbalance is a form of distribution in which majority of the data samples are present in head classes and the occurrence of per class data samples decreases exponentially as we move towards tail classes (Fig.~\ref{fig:overview}(\textit{left})). This family of distribution represents natural distribution for species' population~\cite{inat19}, objects~\cite{wang2017learning} etc. As these distributions are natural, a lot of work has been done to learn discriminative models (\ie classifiers)~\cite{cao2019learning, cui2019classbalancedloss, menon2021longtail, kang2019decoupling, yang2020rethinking, zhou2020BBN} which work across all classes, despite training data following a long-tailed distribution. However, even though there has been a lot of interest, there are still only a handful works which aim to learn generative models for long-tailed distribution. Mullick \etal~\cite{Mullick_2019_ICCV} developed GAMO which aims to learn how to oversample using a generative model. Class Balancing GAN (CBGAN) with a Classifier in the Loop~\cite{rangwani2021class} is the only work which aims to learn a GAN to generate good quality images across classes (in a long-tailed distribution). However, their model is an unconditional GAN which requires a trained classifier to guide the GAN. The requirement of such a classifier can be restrictive. In this work we aim to develop conditional GANs which use class labels, does not require external classifier and generate good quality images (even for tail classes) when trained using long-tailed training data.

\section{Approach}
\begin{figure*}[t]    
    \centering
    \includegraphics[width=\textwidth]{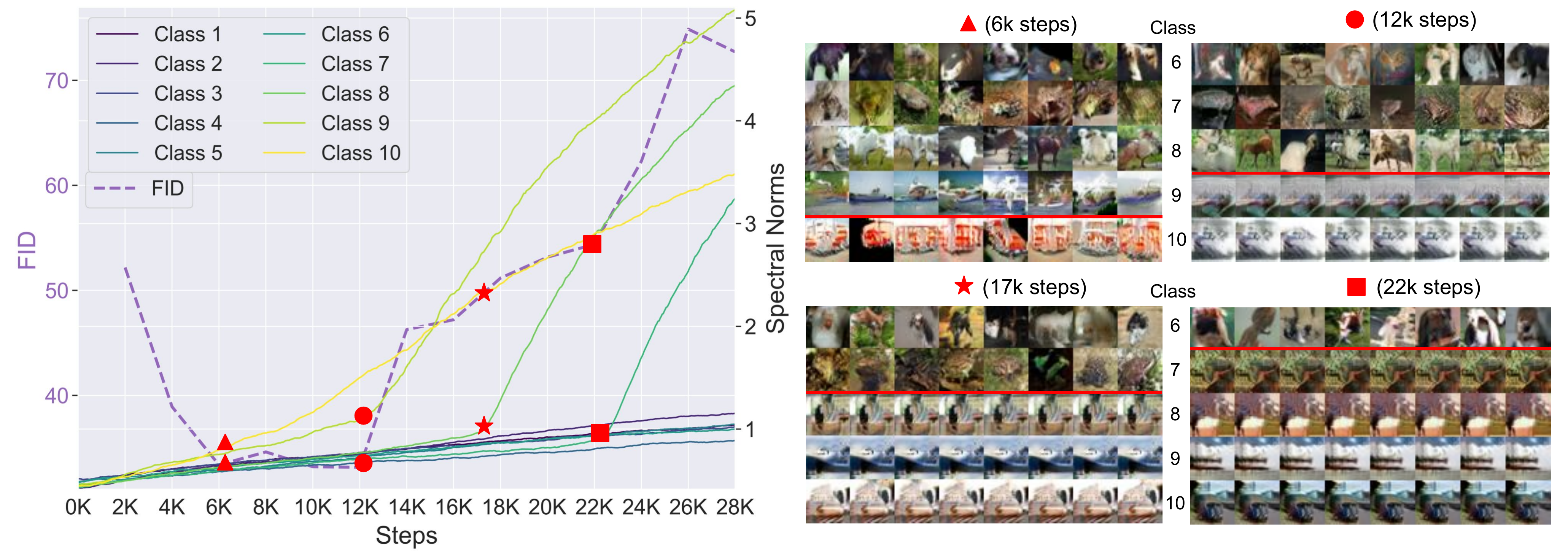}
    \caption{\textbf{Correlation between class-specific mode collapse and spectral explosion.} \emph{(left)} FID/Spectral Norms of class-specific gain parameter of conditional BatchNorm layer on CIFAR-10. \textcolor{red}{Symbols} on plot indicate that FID score's increase correlates with onset of spectral explosion on 4 \textcolor{brown}{tail classes} respectively. \emph{(right)} Images generated for \textcolor{brown}{tail classes} at these train steps reveals corresponding class-specific mode collapse. 
    \label{fig:sn_fid}
    }
\end{figure*}
We start by describing conditional Generative Adversarial Networks (Sec. \ref{subsec:GAN}) and the associated class-specific mode collapse (Sec. \ref{subsec:class-specific-collapse}). Following that we introduce our regularizer formulation, and explain the decorrelation among features caused by gSR that mitigates the mode collapse for tail classes (Sec. \ref{sec:regularizer}).
\subsection{Conditional Generative Adversarial Networks}
\label{subsec:GAN}
Generative Adversarial Networks (GAN) are a combination of two networks the generator $G$ and discriminator $D$. The discriminator's goal is to classify images from training data distribution ($P_{r}$) as real and the images generated through $G$ as fake. In a conditional GAN (cGAN), the generator and discriminator are also enriched with the information about the class label $y$ associated with the image $\mathbf{x} \in \mathbb{R}^{3\times H \times W}$. The conditioning information $y$ helps the cGAN in generating diverse images of superior quality, in comparison to unconditional GAN. The cGAN objectives can be described as:
\begin{equation}
\begin{split}
      &\max_D V(D) = \underset{x \sim P_{r}}{\mathbb{E}}[f_\mathcal{D} D(x|y)] + \underset{z \sim P_{z}}{\mathbb{E}}[f_{\mathcal{G}}(1 - D(G(z|y)))] \\
     & \min_G \mathcal{L}_{G} = \underset{z \sim P_{z}}{\mathbb{E}}[f_{\mathcal{G}}(1 - D(G(z|y)))] 
\end{split}
\end{equation}
where $p_z$ is the prior distribution of latents, $f_{\mathcal{D}}, f_{\mathcal{G}}$, and $g_\mathcal{{G}}$ refer to the mapping functions from which different formulations of GAN can be derived (ref.~\cite{liu2021generative}). The generator $G(z|y)$ generates images corresponding to the class label $y$. In earlier works the conditioning information $y$ was concatenated with the noise vector $z$, however recently conditioning each layer using cBN layer has shown improved performance~\cite{miyato2018cgans}. For a feature $\mathbf{x^l_y} \in \mathbb{R}^d$ conditioned on class $y$ (out of $K$ classes) corresponding to layer $l$, the transformation can be described as:
\begin{equation}
    \mathbf{\hat{x}_{y}^{l}} = \frac{\mathbf{{x}_{y}^{l}} - \mathbf{\mu_B^{l}}}{\sqrt{{\mathbf{\sigma^{l}_B}}^2  + \epsilon}} \rightarrow \mathbf{\gamma^l_{y}} \mathbf{\hat{x}_{y}^{l}} + \mathbf{\beta^l_{y}}
\end{equation}

The $\mathbf{\mu_B^{l}}$ and ${\mathbf{\sigma^{l}_B}}^2$ are the mean and the variance of the batch respectively. The $\mathbf{\gamma^l_{y}} \in \mathbb{R}^d$ and the $\mathbf{\beta^l_{y}} \in \mathbb{R}^d$ are the cBN parameters which enable generation of the image for specific class $y$. We focus on the behaviour of these parameters in the subsequent sections.

\subsection{Class-Specific Mode Collapse}
\label{subsec:class-specific-collapse}
Due to widespread use of conditional GANs~\cite{brock2018large, miyato2018cgans}, it is important that these models are able to learn across various kinds of training data distributions.
However, while training a conditional GAN on long-tailed data distribution, we observe that 
GANs suffer from model collapse on tail classes (Fig.~\ref{fig:overview}). This leads to only a single pattern being generated for a particular tail class. To investigate the cause of this phenomenon, we inspect the class-specific parameters of cBN, which are gain $\mathbf{\gamma^l_{y}}$ and bias $\mathbf{\beta^l_{y}}$.
In existing works, characteristics of groups of features have been insightful for analysis of neural networks and have led to development of regularization techniques  \cite{wu2018group, huang2021group}.
Hence for further analysis we also create $n_g$ groups of the $\mathbf{\gamma^y_{l}}$ parameters and stack them to obtain $\mathbf{\Gamma^l_y} \in \mathbb{R}^{n_g \times n_c}$, where $n_c$ are the number of columns after grouping. It is observed that the value of spectral norm ($\sigma_{\max}(\mathbf{\Gamma_y^l}) \in \mathbb{R}$) explodes (\ie increases abnormally) as mode collapse occurs for corresponding tail class $y$  as shown in Fig.~\ref{fig:sn_fid}. We observe this phenomenon consistently across both the smaller SNGAN~\cite{miyato2018spectral} (Fig. \ref{fig:sn_fid}) and the larger BigGAN~\cite{brock2018large} (Fig.~\ref{fig:overview}) model. We observe similar spectral explosion for BigGAN model as in Fig. \ref{fig:sn_fid} (empirically shown in Fig. \ref{fig:fid_sn}). In the earlier works, mode collapse could be detected by anomalous behaviour of spectral norm of discriminator (refer to suppl. material for details). However in the class-specific mode collapse the discriminator's spectral norms show normal behavior and are unable to signal such collapse. Here, our analysis of $\sigma_{\max}(\mathbf{\Gamma^l_y})$ helps in detecting class-specific mode collapse. \\

\subsection{Group Spectral Regularizer (gSR)}
\label{sec:regularizer}
\begin{figure}[t]
    \centering
    \includegraphics[width=\textwidth]{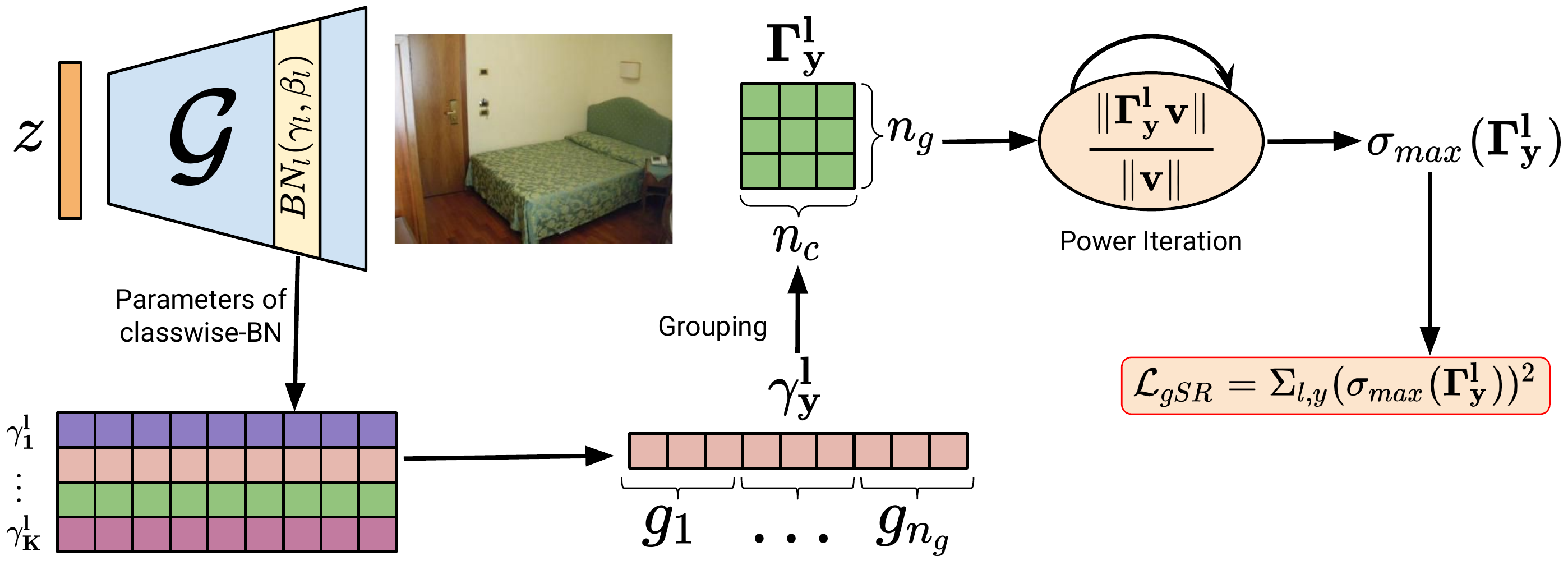}
    \caption{\textbf{Algorithmic overview.} During each training step, 1) we extract the gain $\mathbf{\gamma^l_y}$ for each cBN layers in $G$, 2) group them into matrix $\mathbf{\Gamma^l_y}$ and estimate $\sigma_{\max} (\Gamma^l_y)$. 3) We repeat the same procedure with bias $\mathbf{\beta^l_y}$ to obtain $\sigma_{\max} (\mathbf{B^l_y})$. 4) Finally, we regularize both as described in $\mathcal{L}_{gSR}$ (Eq.~\ref{eq:reg_loss}).}
    \label{fig:algo}
\end{figure}

Our aim now is to prevent the class-specific mode collapse while training. To achieve this, we introduce a regularizer for the generator $G$ which prevents spectral explosion. We would like to emphasize that earlier works including Augmentations~\cite{zhao2020differentiable}, LeCam~\cite{tseng2021regularizing} regularizer etc. are applied on discriminator, hence our regularizer's focus on G is complementary to those of existing techniques. As we observe that spectral norm explodes for the $\mathbf{\gamma^l_{y}}$ and $\mathbf{\beta^l_{y}}$, we deploy a group Spectral Regularizer (gSR) to prevent mode collapse. Steps followed by gSR for estimation of $\sigma_{\max}$ of $\mathbf{\gamma^l_{y}} (\in \mathbb{R}^d)$ are described below (also given in Fig. \ref{fig:algo}):
\begin{equation}
    \label{eq:grouping}
    Grouping: \; \mathbf{\Gamma^l_y} = \Pi(\mathbf{\gamma^l_y}, n_g) \in \mathbb{R}^{n_g \times n_c} 
\end{equation}
\begin{equation}
    \label{eq:power_iter}
    Power\ Iteration: \; \sigma_{\max}(\mathbf{\Gamma^l_y}) = \max_{\mathbf{v}} {\Vert\mathbf{\Gamma^l_yv}\Vert}/{\Vert \mathbf{v}\Vert}
\end{equation}

$\mathbf{\mathbf{v}}(\in \mathbb{R}^d)$ is a random vector for power iterations. $n_g$ and $n_c$ are the number of groups and number of columns respectively, such that $n_g \times n_c = d$ . After estimation of $\sigma_{\max}(\mathbf{\Gamma^l_y})$ and similarly  $\sigma_{\max}(\mathbf{B^l_y})$, the regularized loss objective for generator can be written as:
\begin{equation}
\label{eq:reg_loss}
    \min_{G} \mathcal{L}_{    G} + \lambda_{gSR} \mathcal{L}_{gSR} ; \; \; \text{where} \; \;     \mathcal{L}_{gSR} = \sum_l \sum_{y} \lambda_y (\sigma^2_{\max}(\mathbf{\Gamma_{y}^l}) + \sigma^2_{\max}(\mathbf{B_{y}^l}))
\end{equation}

As the spectral explosion is prominent for the tail classes, we weigh the spectral regularizer term with $\lambda_y$ which has an inverse relation with number of samples $n_y$ in class $y$. Prior work~\cite{cao2019learning} shows that directly using $1/n_y$ can be over-aggressive hence, we use the effective number of samples (a soft version of inverse relation) formally given as (where $\alpha = 0.99$): $\lambda_y = {(1 - \alpha)}/{(1 - \alpha^{n_y})}$.

The regularized objective is used to update weights using backpropagation. Spectral regularizers are used in earlier works~\cite{vahdat2020NVAE, yoshida2017spectral} but they are applied on network weights $W$, whereas to the best of our knowledge, ours is the first work that proposes the regularization of the batch normalization (BatchNorm) parameters. There exist other form of techniques like Spectral Normalization and Spectral Restricted Isometry Property (SRIP)~\cite{bansal2018can} regularizer, which we empirically did not find to be effective for our work (comparison in Sec.~\ref{sec:srsn}). \\

\noindent\textbf{Decorrelation Effect (Relation with Group Whitening):}
\begin{figure*}[t]    
    \centering
    \includegraphics[width=\textwidth]{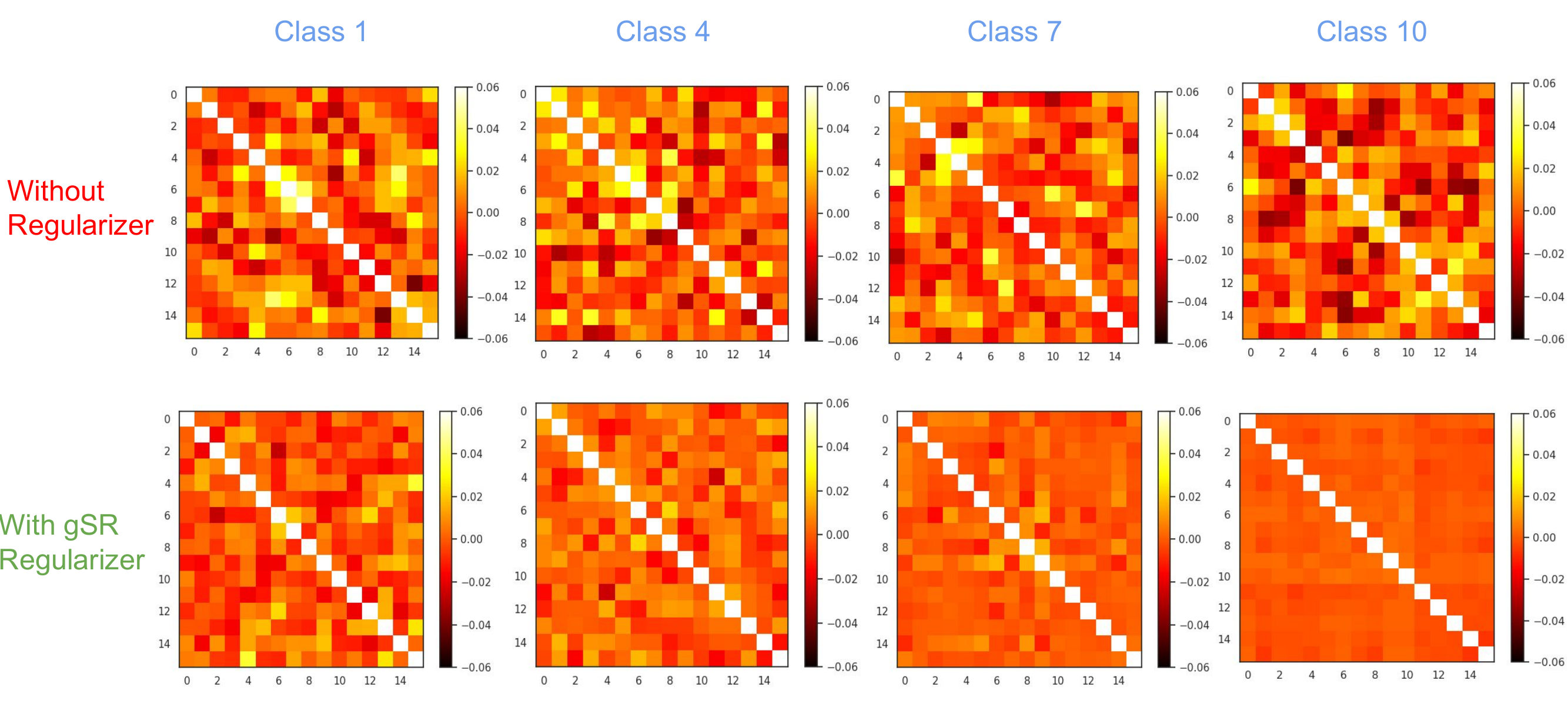}
    \caption{\textbf{Covariance matrices of $\mathbf{\Gamma_y^l}$ for (l = 1) for SNGAN baseline.}  After using gSR (for tail classes with high $\lambda$) the covariance matrix converges to a diagonal matrix in comparison to without gSR (where large correlations exist). This demonstrates the decorrelation effect of gSR on $\mathbf{\gamma_y^l}$, which alleviates class-specific mode collapse.}
    \label{fig:corr_plot}
\end{figure*}
\label{sec:modecollapse}
Group Whitening~\cite{huang2021group} is a recent work which whitens the activation map $X$ by grouping, normalizing and whitening using Zero-phase Component Analysis (ZCA) \nolinebreak{to obtain $\hat{X}$.} 
Due to whitening, the rows of $\hat{X_g}$ get decorrelated, which can be verified by finding the similarity of covariance matrix $\frac{1}{n_c}\hat{X_g}\hat{X_g}^{\intercal}$ to a diagonal matrix. The Group Whitening transformation significantly improves the generalization performance by learning diverse features. As our regularizer also operates on groups of features, we find that minimizing the $\mathcal{L}_{gSR}$ loss also leads to decorrelation of the rows of $\mathbf{\Gamma_y^l}$. We verify this phenomenon by visualizing the covariance matrix $\frac{1}{n_c}[\mathbf{\Gamma_y^l} - \mathbb{E}[\mathbf{\Gamma_y^l}]] ([\mathbf{\Gamma_y^l} - \mathbb{E}[\mathbf{\Gamma_y^l}]])^{\intercal}$.

In Fig.~\ref{fig:corr_plot}, we plot the covariance matrices for both the SNGAN and SNGAN with regularizer (gSR). We clearly observe that for tail classes with high $\lambda_y$ the covariance matrix is more closer to a diagonal matrix which confirms the decorrelation of parameters caused by gSR . 
We find that decorrelation is required more in layers with more class-specific information (\ie earlier layers of generator) rather than layers with generic features like edges. We provide the visualizations for more layers in the suppl. material. 

Recent theoretical results~\cite{wang2020mma,jin2020does} for supervised learning show that decorrelation of parameters mitigates overfitting, and leads to better generalization. This is analogous to our observation of decorrelation being able to prevent mode collapse and helpful in generating diverse images.

\section{Experimental Evaluation}
\label{sec:expts}
We perform extensive experiments on various long-tailed datasets with different resolution. For the controlled imbalance setting, we perform experiments on CIFAR-10~\cite{Krizhevsky09learningmultiple} and LSUN~\cite{journals/corr/YuZSSX15}, which are commonly used in the literature~\cite{cao2019learning, cui2019classbalancedloss, santurkar2018classification} (Sec.~\ref{sec:synth_dist_results}). We also show results on challenging real-world datasets (with skewed data distribution) of iNaturalist2019~\cite{inat19} and AnimalFaces~\cite{kolouri2016sliced} (Sec.~\ref{sec:natural_dist_results}). 

\noindent\textbf{Datasets:} We use the CIFAR-10~\cite{Krizhevsky09learningmultiple} and a subset (5 classes) of LSUN dataset~\cite{journals/corr/YuZSSX15} (50k images balanced across classes) for our experiments. 
The choice of 5 class subset is for a direct comparison with related works~\cite{rangwani2021class, santurkar2018classification} which identify this subset as challenging and use that for experiments.
For converting the balanced subset to the long-tailed dataset with imbalance ratio ($\rho$) (\ie ratio of highest frequency class to the lowest frequency class), we remove the additional samples from the training set. Prior works~\cite{cao2019learning, cui2019classbalancedloss, menon2021longtail} follow this standard process to create benchmark long-tailed datasets. We keep the validation sets balanced and unchanged to evaluate the performance by treating all classes equally. We provide additional details about datasets in the suppl. material. We perform experiments on the imbalance ratio of 100 and 1000. In case of CIFAR-10 for $\rho = 1000$ the majority class contains 5000 samples whereas the minority class has only 5 samples. For performing well in this setup, the GAN has to robustly learn from many-shot (5000 sample class) as well as at few-shot (5 sample class) together, making this benchmark challenging.

\setlength{\intextsep}{0pt}%
\begin{table*}[t]
    \centering
    \caption{\textbf{Quantitative results on the CIFAR-10 and LSUN dataset.} On an average, we observe a relative improvement in FID of 20.33\% and 39.08\% over SNGAN and BigGAN baselines respectively.}
    \label{tab:main_results}
    \resizebox{\textwidth}{!}{\begin{tabular}{lcccc|cccc}
    \toprule
    & \multicolumn{4}{c}{CIFAR-10} & \multicolumn{4}{c}{LSUN} \\ \hline
         Imb. Ratio ($\rho$) & \multicolumn{2}{c}{100} & \multicolumn{2}{c}{1000} & \multicolumn{2}{c}{100} &\multicolumn{2}{c}{1000}  \\ \hline
         & FID ($\downarrow$)&      IS($\uparrow$)& FID ($\downarrow$)&      IS($\uparrow$) & FID ($\downarrow$)&      IS($\uparrow$)& FID ($\downarrow$)&      IS($\uparrow$)  \\\midrule

         CBGAN~\cite{rangwani2021class} &33.01$_{\pm0.12}$ &6.58$_{\pm0.05}$ &44.82$_{\pm0.12}$ &5.92$_{\pm0.05}$ &37.41$_{\pm0.10}$ &2.82$_{\pm0.03}$ &44.70$_{\pm0.13}$ &2.77$_{\pm0.02}$\\
         LSGAN~\cite{mao2017least} &24.36$_{\pm0.01}$ &7.77$_{\pm0.07}$ &51.47$_{\pm0.21}$ &6.54$_{\pm0.05}$ &37.64$_{\pm0.05}$ &3.12$_{\pm0.01}$ &41.50$_{\pm0.04}$ &2.74$_{\pm0.02}$\\
         SNGAN~\cite{miyato2018spectral} &30.62$_{\pm0.07}$   &6.80$_{\pm0.02}$ &54.58$_{\pm0.19}$ &6.19$_{\pm0.01}$ &38.17$_{\pm0.02}$ &3.02$_{\pm0.01}$ &38.36$_{\pm0.11}$ &2.99$_{\pm0.01}$\\
         \rowcolor{gray!10} \; + gSR (Ours) &18.58$_{\pm0.10}$ &7.80$_{\pm0.09}$ &48.69$_{\pm0.04}$ &5.92$_{\pm0.01}$ &28.84$_{\pm0.09}$ &3.50$_{\pm0.01}$ &35.76$_{\pm0.05}$ &3.56$_{\pm0.01}$\\ \midrule
         BigGAN~\cite{brock2018large} &19.55$_{\pm0.12}$ &8.80$_{\pm0.09}$ &50.78$_{\pm0.23}$ &6.50$_{\pm0.05}$ &38.65$_{\pm0.05}$ &\textbf{4.02$_{\pm0.01}$} &45.89$_{\pm0.30}$ &3.25$_{\pm0.01}$\\
         \rowcolor{gray!10} \; + gSR (Ours) &\textbf{12.03$_{\pm0.08}$} &\textbf{9.21$_{\pm0.07}$} &\textbf{38.38$_{\pm0.01}$} &\textbf{7.24$_{\pm0.04}$} &\textbf{20.18$_{\pm0.07}$} &3.67$_{\pm0.01}$ &\textbf{24.93$_{\pm0.09}$} &\textbf{3.68$_{\pm0.01}$} \\ \bottomrule
    \end{tabular}}
\end{table*}

\noindent\textbf{Evaluation: } We report the standard Inception Score (IS)~\cite{salimans2016improved} and Fr\'echet Inception Distance (FID) metrics for the generated datasets. We report the mean and standard deviation of 3 evaluation runs similar to the protocol followed by DiffAug~\cite{zhao2020differentiable} and LeCam~\cite{tseng2021regularizing}. We use a held out set of 10k images for calculation of FID for both the datasets. The held out sets are balanced across classes for fair evaluation of each class. 

\noindent\textbf{Configuration:} We perform experiments by using PyTorch-StudioGAN implemented  by Kang \etal~\cite{kang2020contrastive}, which serves as the baseline for our framework. We generate 32 $\times$ 32 sized images for the CIFAR-10 dataset and 64 $\times$ 64 sized images for the LSUN dataset. For the CIFAR-10 experiments, we use 5 $D$ steps for each $G$ step. Unless explicitly stated, we by default add the following two SOTA regularizers to obtain strong generic baselines for all the experiments (except CBGAN for which we follow exact setup described by Rangwani \etal~\cite{rangwani2021class}):
\begin{itemize}
\itemsep0em
    \item {DiffAugment~\cite{zhao2020differentiable}:} We apply the differential augmentation technique with the colorization, translation, and cutout augmentations.
    \item {LeCam~\cite{tseng2021regularizing}}: LeCam regularizer prevents divergence of discriminator $D$ by constraining its output through a regularizer term $R_{LC}$ (ref. suppl. material).
\end{itemize}
Any improvement over these strong regularizers published recently is meaningful and shows the effectiveness of the proposed methods. We use a batch size of 64 for all our CIFAR-10 and LSUN experiments. For sanity check of the implementation we run the experiments for the balanced dataset (CIFAR-10) case where our FID is similar to the one obtained in LeCam~\cite{tseng2021regularizing}, details are in the suppl. material.

\noindent\textbf{Baselines:} We compare our regularizer with the recent work of Class Balancing GAN (CBGAN) ~\cite{rangwani2021class} which uses an auxiliary classifier for long-tailed image generation. We use the public codebase and configuration provided by the authors.
The auxiliary classifiers are obtained using the LDAM-DRW as suggested by CBGAN authors. We use the SNGAN~\cite{miyato2018spectral} (with projection discriminator~\cite{miyato2018cgans}) as our base method on which we apply the Augmentation and LeCam regularizer for a strong baseline. We also compare our method with LSGAN~\cite{mao2017least}, which is shown to be effective in preventing the mode-collapse (we use the same configuration as in SNGAN for fairness of comparison). To demonstrate improvement over large scale GANs we also use BigGAN \cite{brock2018large} with LeCam and DiffAug regularizers as baseline.  We then add our group Spectral Regularizer (gSR) in the loss terms for BigGAN and SNGAN, and report the performance in \mbox{Table~\ref{tab:main_results}.} We do not use ACGAN as our baseline as it leads to image generation which doesn't match the conditioned class label (\ie class confusion)~\cite{rangwani2021class}.

\subsection{Results on CIFAR-10 and LSUN}
\label{sec:synth_dist_results}

\begin{figure*}[t]
  \centering
\begin{minipage}[c]{0.34\linewidth}
    \centering
    \includegraphics[width=\textwidth]{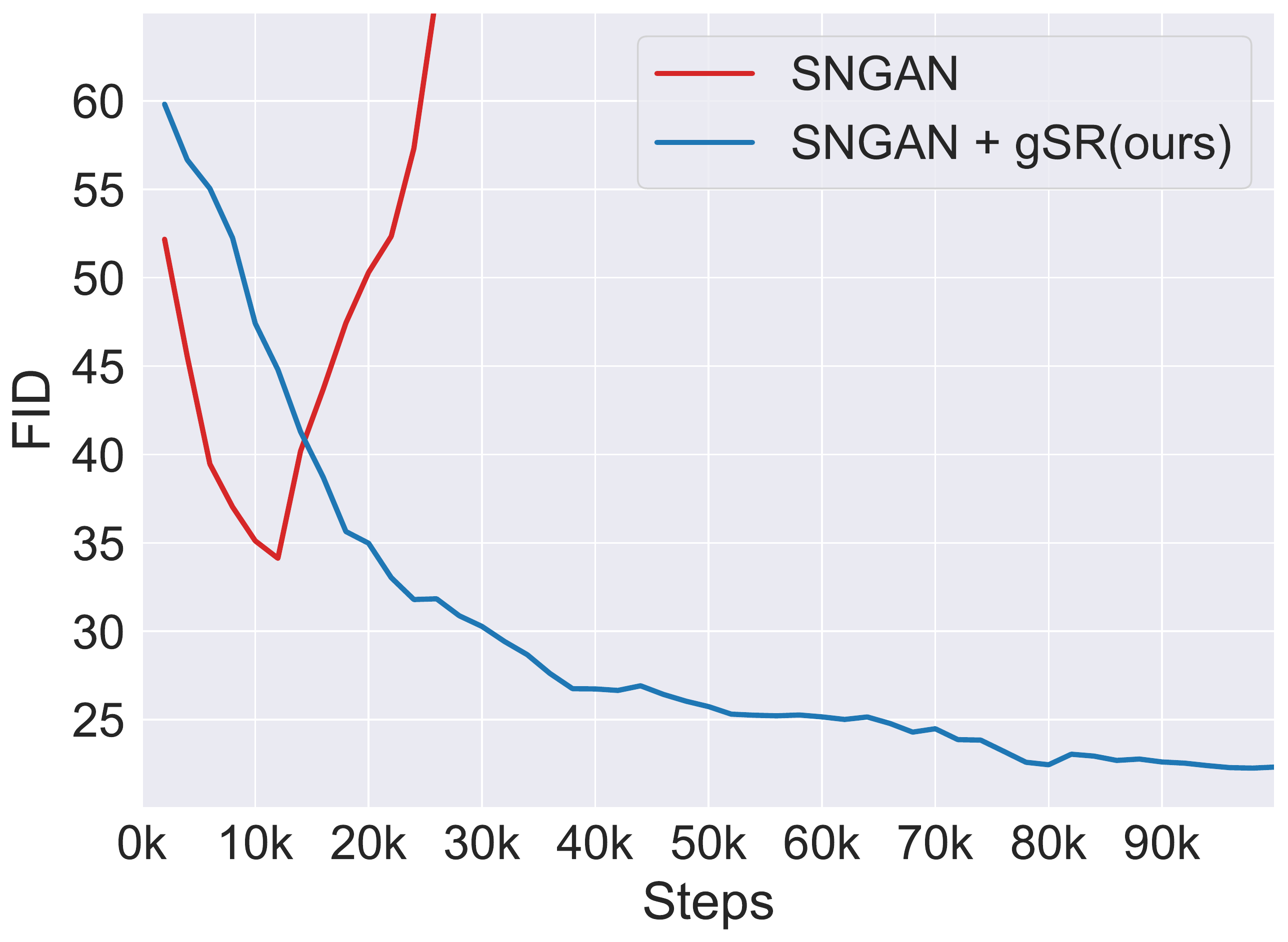}
    \caption{\textbf{Stability}. Addition of gSR (to baseline) stabilizes the training to continually improve, as indicated by the FID scores.}
    \label{fig:fid_gsr}
  \end{minipage}
   \hfill
  \begin{minipage}[c]{0.64\textwidth}
    \centering
    \includegraphics[width=\textwidth]{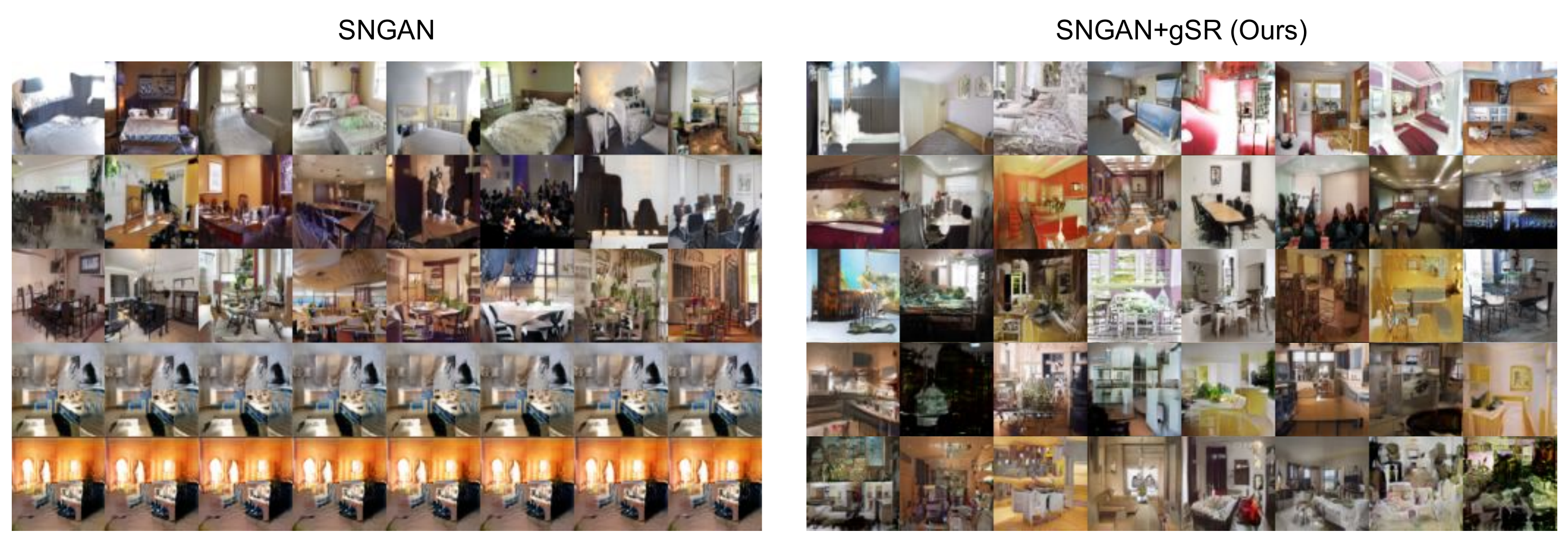}
    \begin{minipage}{1cm}
    \vfill
    \end{minipage}
    \caption{\textbf{Qualitative evaluations of SNGAN baseline on LSUN dataset.} Each row represents images from a class. Note the class-specific mode collapse observed in tail-classes in SNGAN (last two rows), which is alleviated after addition of gSR to generate diverse images.}
    \label{fig:sngan_imgs}
  \end{minipage}
\end{figure*}

\noindent\textbf{Stability:} Fig. \ref{fig:fid_gsr} shows the FID vs iteration steps for the SNGAN and 
SNGAN +gSR configuration. Using gSR regularizer with SNGAN is able to effectively prevent the class-specific mode collapse, which in turn helps the GAN to improve for a long duration of time. SNGAN without gSR starts degrading quickly and stops improving very soon, this shows the stabilizing effect imparted by gSR regularizer in the training process. The stabilizing effect is similarly observed even for the BigGAN (ref. FID plot in Fig.~\ref{fig:overview}).

\noindent\textbf{Comparison of Quality}: We observe that application of regularizer effectively avoids mode collapse and leads to a large average improvement (of 7.46) in FID for the (SNGAN + gSR) case, in comparison to SNGAN baseline across the four datasets (Table \ref{tab:main_results}). Our method is also effective on BigGAN where it is able to achieve SOTA FID and IS significant improvement in almost all cases. Although SNGAN and BigGAN baselines are already enriched with SOTA regularizers of (LeCam + DiffAug) to improve results, yet the addition of our gSR regularizer significantly boosts performance by harmonizing with other regularizers. It also shows that our regularizer complements the existing work and effectively reduces mode collapse. Fig.~\ref{fig:sngan_imgs} shows a comparison of the generated images for the different methods, where gSR is able to produce better images for the tail classes for LSUN dataset (refer Fig.~\ref{fig:overview} for qualitative comparison on CIFAR-10 ($\rho$ = 100)). To quantify improvement over each class, we compute class-wise FID and mean FID (\ie Intra FID) in Fig. \ref{fig:intra-fid}. We find that gSR leads to very significant improvement in tail class FID as it prevents the collapse. Due to the stabilizing effect of gSR we find that head class FID are also improved, clearly demonstrating the benefit of gSR for all classes. We also provide additional metrics (precision~\cite{kynkaanniemi2019improved}, recall~\cite{kynkaanniemi2019improved}, density~\cite{ferjad2020icml}, coverage~\cite{ferjad2020icml} and Intra-FID) in suppl. material. We find that almost all metrics show similar improvement as seen in FID (Table \ref{tab:main_results}).

\subsection{Results on Naturally Occurring Distributions}
\label{sec:natural_dist_results}
\begin{figure*}[t]
  \centering
\begin{minipage}[c]{0.34\linewidth}
    \centering
    \includegraphics[width=\textwidth]{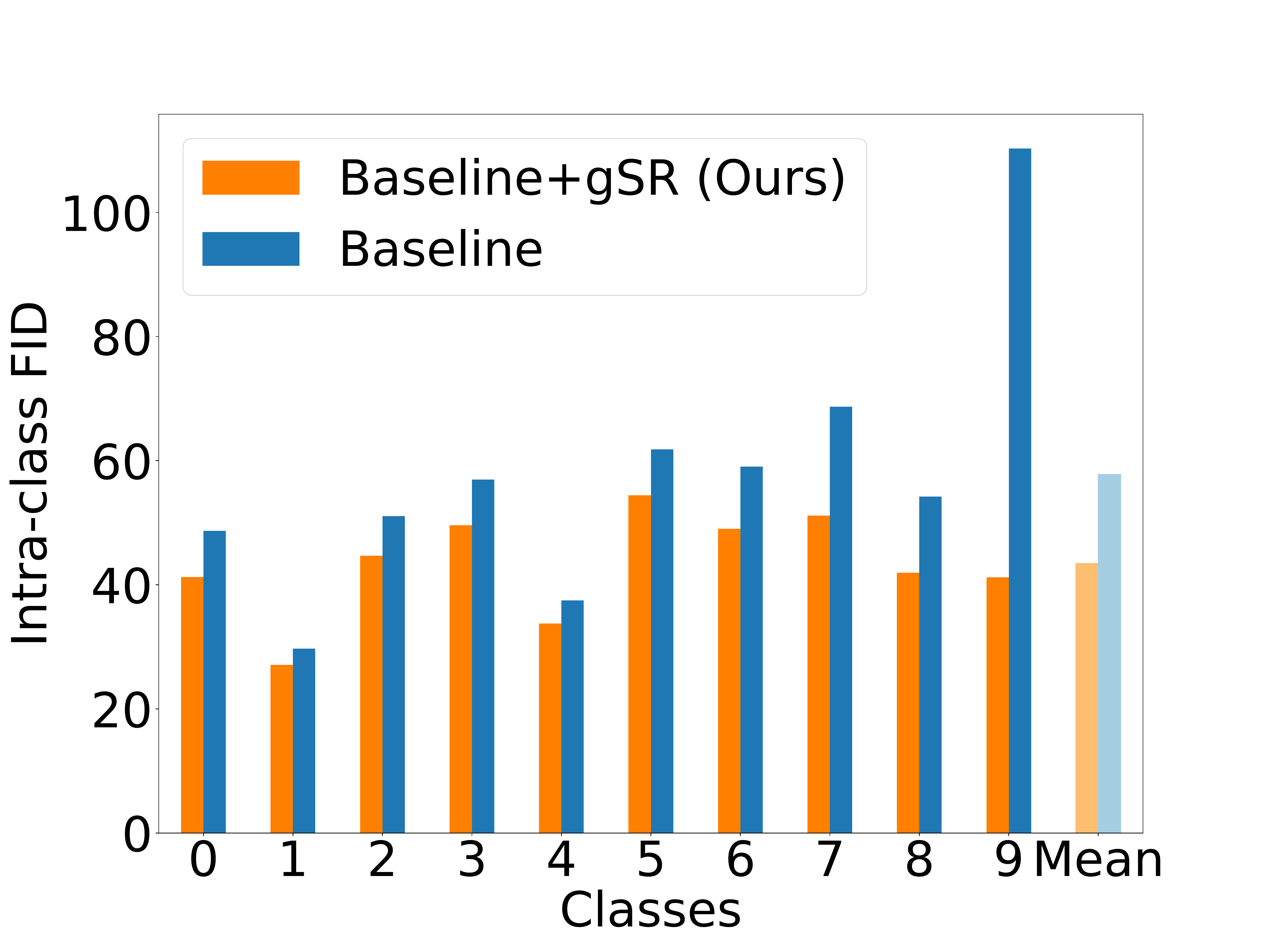}
    \captionof{figure}{Class-Wise FID and mean FID (Intra-FID) of BigGAN on CIFAR-10 over 5$K$ generated images($\rho$ = 100).}
    \label{fig:intra-fid}
  \end{minipage}
   \hfill
  \begin{minipage}[c]{0.60\textwidth}
    \centering
    \captionof{table}{\textbf{Quantitative results on iNaturalist-2019 and AnimalFace Dataset.} We compare mean FID ($\downarrow$) with other existing \mbox{approaches.}}
    \label{tab:iNaturalist}
    \begin{tabular}{l|c|c|c}
    \toprule
                & \multicolumn{2}{c}{iNaturalist 2019} & AnimalFace \\ \hline
         Method & cGAN & FID($\downarrow$) & FID($\downarrow$)\\ \midrule
         SNResGAN$^*$~\cite{miyato2018spectral}& \xmark & 13.03$_{\pm0.07}$ & -\\
         CBGAN$^*$~\cite{rangwani2021class} & \xmark & 9.01$_{\pm0.08}$ & - \\ 
         ACGAN$^*$~\cite{odena2017conditional} & \cmark & 47.15$_{\pm0.11}$ & - \\
         SNGAN$^*$~\cite{miyato2018cgans} & \cmark & 
         21.53$_{\pm0.03}$ & -\\  \midrule
         
         BigGAN~\cite{brock2018large} & \cmark & 6.87$_{\pm0.04}$ & 38.41$_{\pm0.04}$\\
        \rowcolor{gray!10} \; + gSR (Ours) & \cmark&\textbf{6.36$_{\pm0.04}$} & \textbf{35.76$_{\pm0.04}$} \\ \bottomrule
    \end{tabular}
  \end{minipage}
\end{figure*}

To show the effectiveness of our regularizer on natural distributions we perform experiments on two challenging datasets: iNaturalist-2019~\cite{inat19} and AnimalFace~\cite{si2011learning}. %
The iNaturalist dataset is a real-world long-tailed dataset with 1010 
classes of different species. There is high diversity among the images of each class, due to their distinct sources of origin. The dataset follows a long-tailed distribution with around 260k images. The second dataset we experiment with is the Animal Face Dataset~\cite{kolouri2016sliced} which contains 20 classes with with around 110 samples per class. We generate 64 $\times$ 64 images for both datasets using the BigGAN with a batch size of 256 for iNaturalist and 64 for AnimalFaces. The BigGAN baseline is augmented with LeCam and DiffAug regularizers. We compare our method with the baselines described in~\cite{rangwani2021class}. We evaluate each model using the FID on a training subset which is balanced across classes. For baselines we directly report results from Rangwani \etal \cite{rangwani2021class} (indicated by $^*$) in 
Table~\ref{tab:iNaturalist}.

The BigGAN baseline achieves an FID of 6.87 on iNaturalist 2019 dataset, which improves relatively by 7.42\% (-0.51 FID) when proposed gSR is combined with BigGAN. Our approach is also able to achieve FID better than SOTA CBGAN on iNaturalist 2019 dataset.
\mbox{Table~\ref{tab:iNaturalist}} shows the performance of the BigGAN baseline over the AnimalFace dataset, where after combining with our gSR regularizer we see FID improvement by 6.90\% (-2.65 FID). The improvements on both the large long-tailed dataset and few shot dataset of AnimalFace shows that gSR is able to effectively improve performance on real-world datasets. We provide additional experimental details and results in the suppl. material.
\setlength{\intextsep}{0pt}%

\section{Analysis}
\subsection{Ablation over Regularizers}
We use the combination of existing regularizers (LeCam + DiffAug) with our regularizer (gSR) to obtain the best performing models. For further analysis of importance of each, we study their effect in comparison to gSR in this section. We perform experiments by independently applying each of them on vanilla SNGAN. Table~\ref{tab:sngan_abl} shows that existing regularizer in itself is not able to effectively reduce FID, whereas gSR is effectively able to reduce FID independently by 3.8 points. However, we find that existing regularizers along with proposed gSR, make an effective combination which further reduces FID significantly (by 9.27) on long-tailed CIFAR-10 ($\rho = 100$). This clearly shows that our regularizer effectively complements the existing regularizers.
\subsection{High Resolution Image Generation} 
\begin{figure*}[t]    
    \centering
    \includegraphics[width=\textwidth]{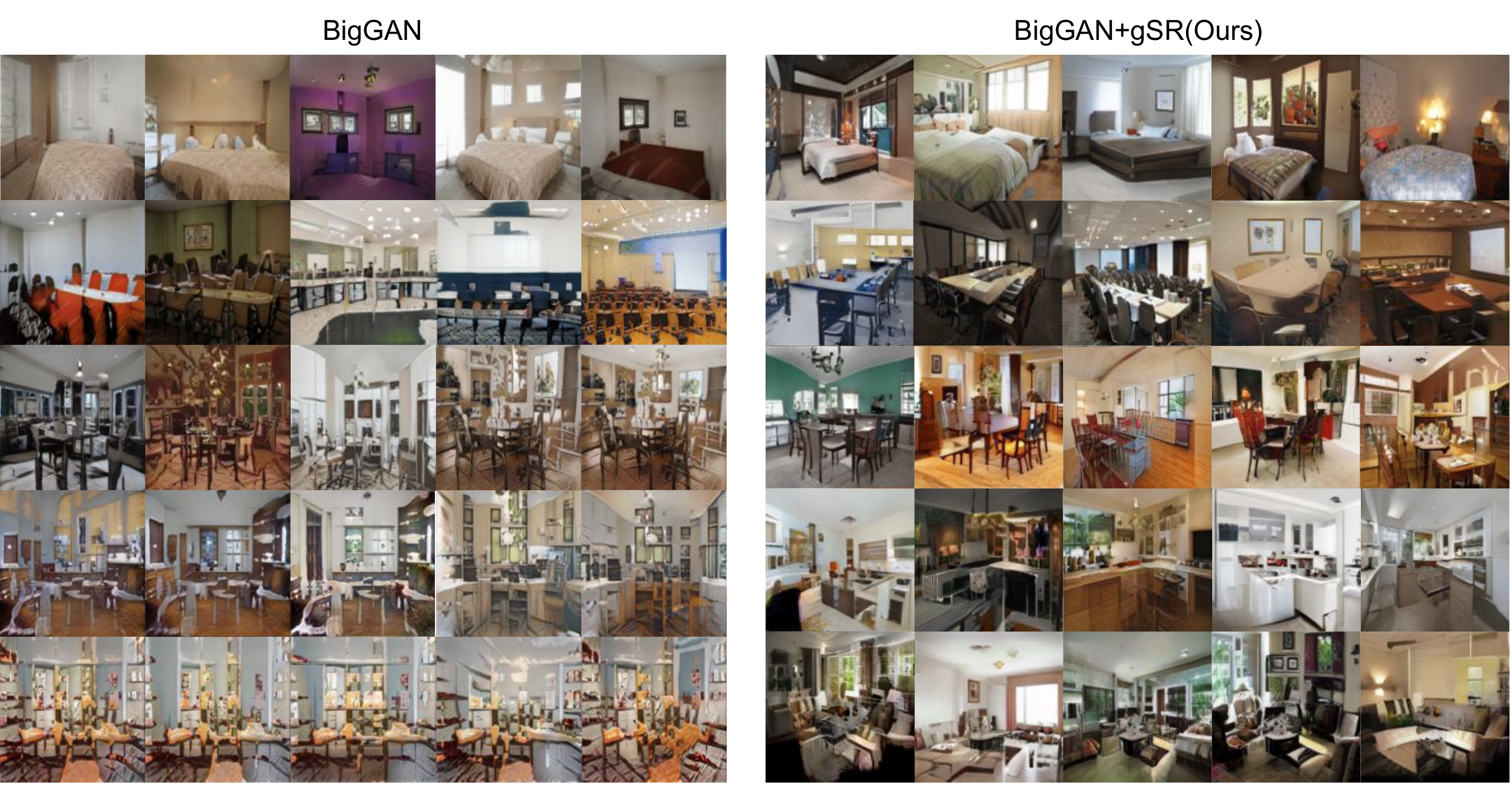}
    \caption{\textbf{Qualitative comparison of BigGAN variants on LSUN dataset ($\rho$=100) (128 $\times$ 128).} Each row represents images from a distinct class. }
    \label{fig:lsun_128}
\end{figure*}

As the LSUN dataset is composed of high resolution scenes we also investigate if the class-specific mode collapse phenomenon when GANs are trained for high resolution image synthesis. For this we train SNGAN and BigGAN baselines for (128 $\times$ 128) using the DiffAugment and LeCAM regularizer (details in suppl. material). We find that similar phenomenon of spectral explosion leading to class-specific collapse occurs (as in 64 $\times$ 64 case), which is mitigated when the proposed gSR regularizer is combined with the baselines (Fig. \ref{fig:lsun_128}). The gSR regularizer leads to significant improvement in FID (Table \ref{tab:high_res}) also seen in qualitatively in Fig. \ref{fig:lsun_128}.

\begin{table*}[t]
\parbox{.48\linewidth}{
\centering
    \caption{\textbf{Ablation over regularizers on SNGAN.} We report FID and IS on  CIFAR-10 dataset (with $\rho = 100$).}
    \label{tab:sngan_abl}
    \begin{tabular}{c|c|c|c}
    \toprule
          \begin{tabular}[c]{@{}l@{}}LeCam+\\DiffAug\end{tabular}  & gSR  & FID($\downarrow$) & IS($\uparrow$)\\ \midrule
            \xmark         &      \xmark     &31.73$_{\pm0.08}$                 &7.18$_{\pm0.02}$      \\  
                        \rowcolor{gray!10}  \xmark         &      \cmark     &27.85$_{\pm0.05}$                  &7.09$_{\pm0.02}$    \\
                        \cmark         &      \xmark     &30.62$_{\pm0.07}$                  &6.80$_{\pm0.02}$      \\   
          \rowcolor{gray!10} \cmark        &      \cmark     &\textbf{18.58$_{\pm0.10}$} &\textbf{7.80$_{\pm0.09}$}                      \\ \bottomrule 
    \end{tabular}
}
\hfill
\parbox{.48\linewidth}{
    \centering
    \caption{\textbf{Image Generation (128 $\times$ 128)}. We report FID on both SNGAN and BigGAN on LSUN dataset (for $\rho$ = 100 and $\rho$ = 1000).}
    \label{tab:high_res}
    
    {\begin{tabular}{lcc}
    \toprule
         Imb. Ratio ($\rho$)  & \multicolumn{1}{c}{100} &\multicolumn{1}{c}{1000}  \\ \hline
         &  FID ($\downarrow$)& FID ($\downarrow$) \\\midrule
         
         SNGAN~\cite{miyato2018spectral} &53.91$_{\pm0.02}$ &72.37$_{\pm0.08}$\\
         \rowcolor{gray!10} \; + gSR (Ours)  &\textbf{25.31$_{\pm0.03}$} &\textbf{31.86$_{\pm0.03}$}\\ \midrule
         BigGAN~\cite{brock2018large}   &61.63$_{\pm0.11}$ &77.17$_{\pm0.18}$\\
         \rowcolor{gray!10} \; + gSR (Ours)   &\textbf{16.56$_{\pm0.02}$} &\textbf{45.08$_{\pm0.10}$} \\ \bottomrule
    \end{tabular}
}}
\end{table*}

\subsection{Comparison with related Spectral Regularization and Normalization Techniques}
\label{sec:srsn}
\begin{figure*}[t]
  \centering
  \begin{minipage}[t]{0.45\linewidth}
    \centering
    \includegraphics[width=1\textwidth, height=2.5cm]{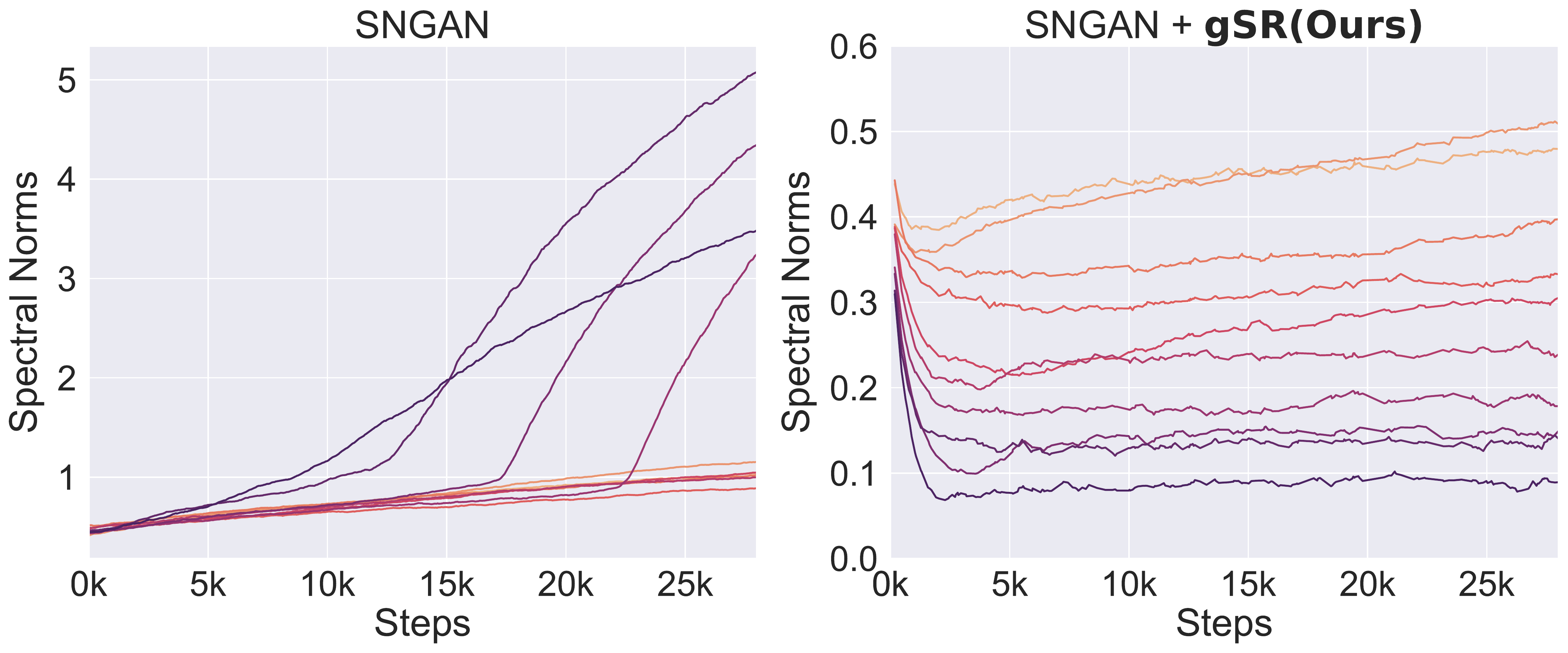}
  \end{minipage}
  \begin{minipage}[t]{0.54\linewidth}
    \centering
    \includegraphics[width=1\textwidth,height=2.5cm]{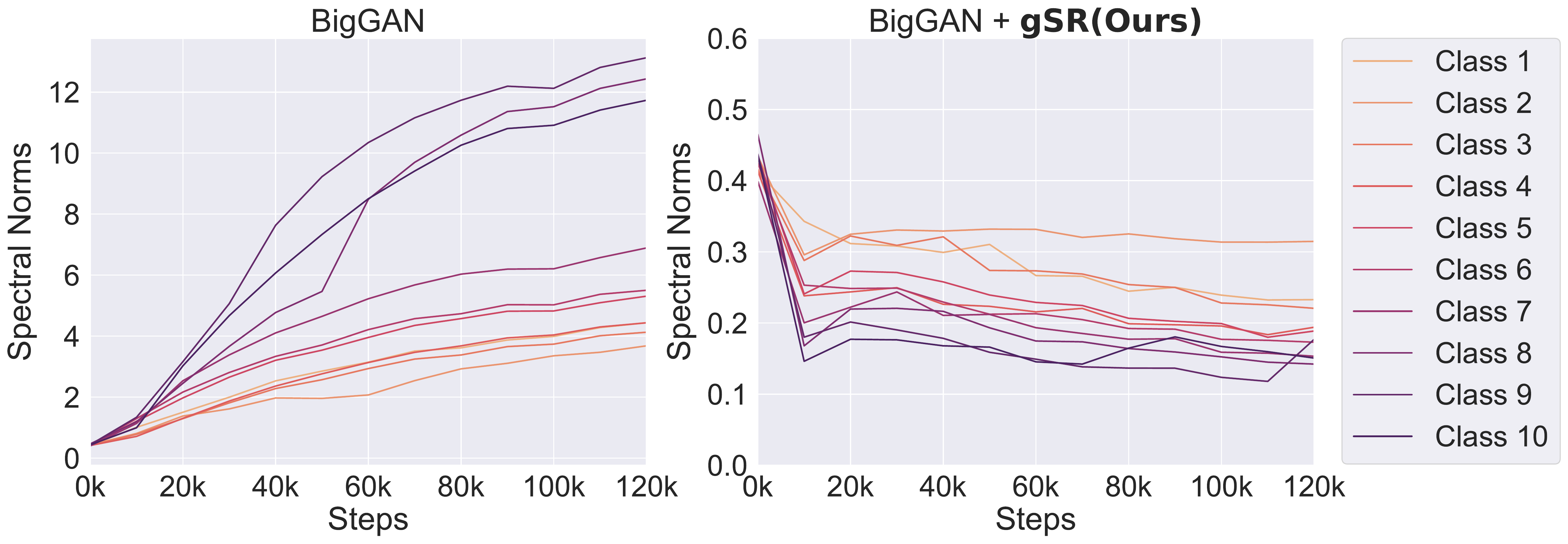}
  \end{minipage}
    \caption{\textbf{Effect of gSR on spectral norms of $\mathbf{\Gamma^l_y}$ (CIFAR-10).} We observe a spectral explosion both for SNGAN(\textit{left}) and BigGAN(\textit{right}) baselines of tail classes' cBN parameters. This is prevented by addition of gSR as shown on corresponding right.}
  \label{fig:fid_sn}
\end{figure*}

As gSR constrains the exploding spectral norms for the cBN parameters (Fig.~\ref{fig:fid_sn})
to evaluate its effectiveness, we test it against other variants of spectral normalization and regularization techniques on SGAN for CIFAR-10 ($\rho=100$).

\noindent\textbf{Group Spectral Normalization (gSN) of BatchNorm Parameters:} In this setting, rather than using sum of spectral norms (Eq. \ref{eq:reg_loss}) as regularizer for the class-specific parameters of cBN in gSR, we normalize them by dividing it by group spectral norms (\ie $\frac{\mathbf{\gamma^l_y}}{\sigma_{max}(\mathbf{\Gamma^l_y})}$) \cite{miyato2018spectral}.

\noindent\textbf{Group Spectral Restricted Isometry Property (gSRIP) Regularization:}  Extending SRIP~\cite{bansal2018can}, the class-specific parameters of cBN which are grouped to form a matrix $\mathbf{\Gamma^l_y}$, the regularizer is the sum of square of spectral norms of $(\mathbf{\Gamma^l_y}^\intercal\mathbf{\Gamma^l_y} - \mathbf{I)}$, (instead that of $\mathbf{\Gamma^l_y}$ in gSR (Eq. \ref{eq:reg_loss})). We report our findings in Table~\ref{tab:compare_reg_norm}. It can be inferred that all three techniques, namely gSN, gSRIP, and gSR, lead to significant improvements over the baseline. This also confirms our hypothesis that reducing (or constraining) spectral norm of cBN parameters alleviates class-specific mode collapse. However, it is noteworthy that gSR gives the highest boost over the baseline by a considerable margin in terms of FID.

\subsection{Analysis of gSR}
In this section (and suppl. material) we provide ablations of gSR using long-tailed CIFAR-10 ($\rho$=100).  

 \noindent \textbf{Can gSR work with StyleGAN-2?} We train and analyze the StyleGAN2-ADA implementation available~\cite{kang2020contrastive} on long-tailed datasets, where we find it also suffers from class-specific mode collapse.
 \setlength{\intextsep}{0pt}%
  \begin{wrapfigure}{r}{0.4\textwidth}
    \includegraphics[width=0.4\textwidth]{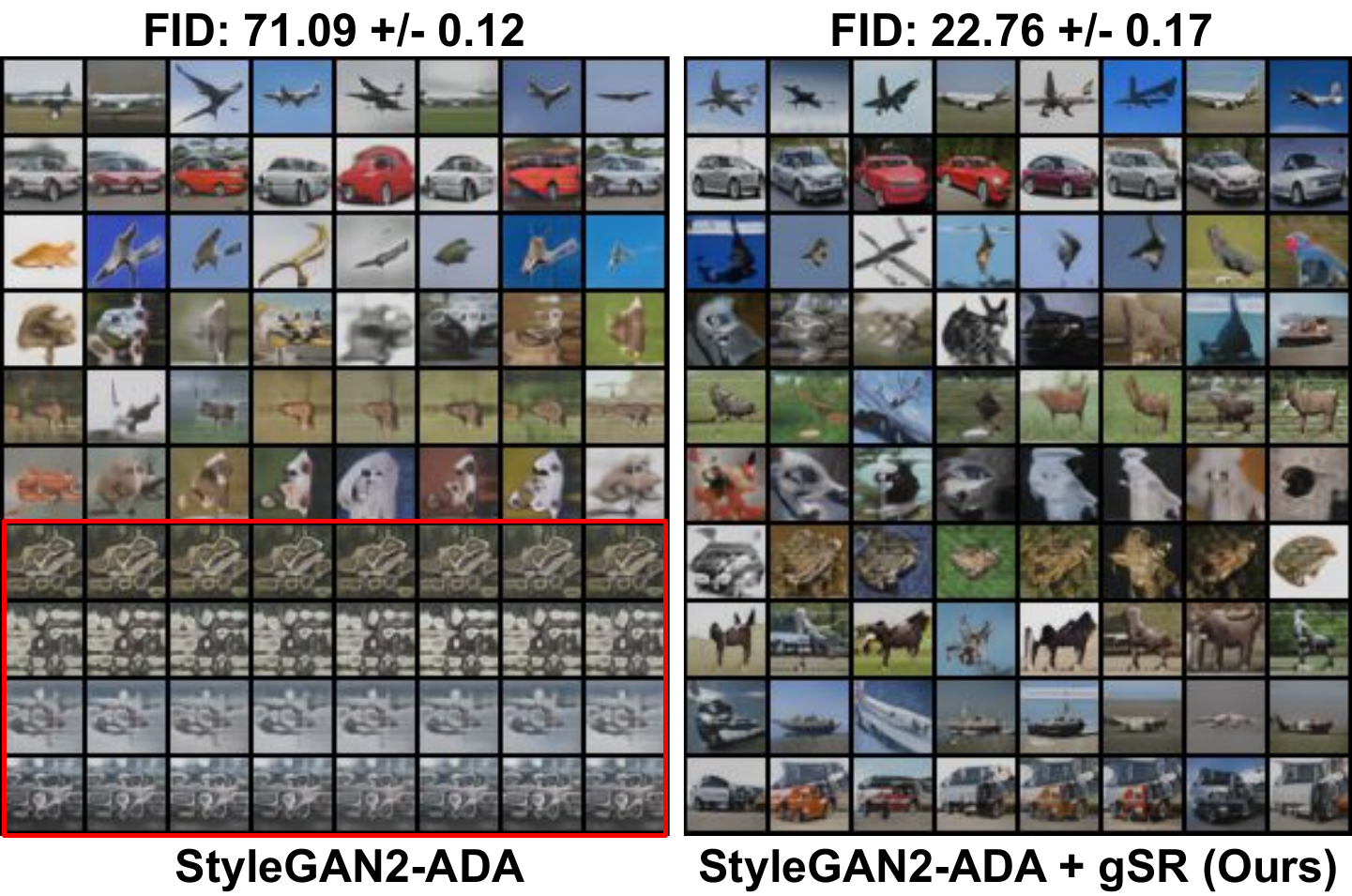}
    \caption{\textbf{StyleGAN2-ADA}  On CIFAR-10 ($\rho = 100$), comparison of gSR with the baseline.}
    \label{fig:qual_results}
\end{wrapfigure} We then implement gSR for StyleGAN2 by grouping 512 dimensional class conditional embeddings in mapping network to 16x32 and calculating their spectral norm which is added to loss (Eq. \ref{eq:reg_loss}) as $R_{gSR}$.We find that gSR is able to effectively prevent the mode collapse (Fig.\ \ref{fig:qual_results}) and also results in significant improvement in FID in comparison to StyleGAN2-ADA baseline. Further analysis and results are present in suppl. material.
 
\noindent\textbf{What is gSR's effect on spectral norms?}  We plot spectral norms of class-specific gain parameter of 1$^{st}$ layer of generator in SNGAN. Spectral norms explode for the tail classes without gSR, while they remain stable when gSR is used. Fig.~\ref{fig:fid_gsr} for same experiment shows that while using gSR the FID keeps improving, whereas it collapses without using gSR.  This confirms our hypothesis that constraining spectral norms stabilizes the training. We find that a similar phenomenon also occurs for BigGAN (Fig. ~\ref{fig:fid_gsr}) which uses SN in $G$, which shows that gSR is complementary to SN.

\noindent\textbf{What should be the ideal number of groups?} Grouping of the ($\mathbf{\gamma^l_y}$) into $\mathbf{\Gamma^l_y} \in \mathbb{R}^{{n_g}\times{n_c}}$ is a central operation in our regularizer formulation (Eq. \ref{eq:grouping}). We group $\mathbf{\gamma^l_y}$ (and $\mathbf{\beta^l_y}$) $\in \mathbb{R}^{256}$ into a matrix $\mathbf{\Gamma^l_y}$ (and $\mathbf{B^l_y}$) ablate over different combinations of $n_g$ and $n_c$. Table \ref{tab:group_ablations} shows that FID scores do not change much significantly with $n_g$. As we also use power iteration to estimate the spectral norm $\sigma_{\max} (\mathbf{\Gamma^l_y})$, we report iteration complexity (multiplications). Since grouping into square matrix($n_g = 16$) gives slightly better FIDs while also being time efficient we use it for our experiments. We also provide additional mathematical intuition for the optimality of choice of $n_c = n_g$ in suppl. material.

\begin{table*}[t]
\parbox{.4\linewidth}{
\centering
    \centering
    \caption{\textbf{Quantitative comparison of spectral regularizers.} Comparison against Different Spectral Norm Regularizers on grouped cBN parameters.}
    \label{tab:compare_reg_norm}
    \begin{tabular}{l|cc}
    \toprule
         & FID($\downarrow$) & IS($\uparrow$) \\ \midrule
         SNGAN~\cite{miyato2018cgans} & 30.62$_{\pm0.07}$ & 6.80$_{\pm0.07}$ \\
        + gSN~\cite{miyato2018spectral}  & 23.97$_{\pm0.13}$ & 7.49$_{\pm0.05}$ \\
        + gSRIP~\cite{bansal2018can}  & 23.67$_{\pm0.02}$ & 7.79$_{\pm0.06}$ \\
        \rowcolor{gray!10} + gSR (Ours)  & \textbf{18.58}$_{\pm0.10}$ & \textbf{7.80}$_{\pm0.09}$ \\ \bottomrule
    \end{tabular}
}
\hfill
\parbox{.55\linewidth}{
        \centering
    \caption{\textbf{Group size ablations.} We report average FID and IS on CIFAR-10 dataset. $n_g=16$ gives the best FID while also being computationally efficient, measured by per Iteration (Iter.) complexity. (Iter. complexity for power iteration method is calculated as ($n_g^2$ + $n_c^2$) x number of power iterations (4 in our setting)).}
    \label{tab:group_ablations}
    \resizebox{\linewidth}{!}{
    {\begin{tabular}{c|c|c|c|c} 
    \toprule
    $n_g$ & $n_c$ & FID($\downarrow$) & IS($\uparrow$) & Iter. Complexity($\downarrow$) \\ \midrule
         4 & 64 & 20.16$_{\pm0.03}$ & \textbf{7.96$_{\pm0.01}$} & 16448 \\
         8 & 32 & 18.69$_{\pm0.06}$ & 7.80$_{\pm0.01}$ & 4352\\
         16 & 16 & \textbf{18.58$_{\pm0.10}$} & 7.80$_{\pm0.09}$ & \textbf{2048}\\
         32 & 8 & 20.19$_{\pm0.06}$ & 7.85$_{\pm0.01}$ & 4352\\ \bottomrule
    \end{tabular}}
}}
\end{table*}
\section{Conclusion and Future Work}
In this work we identify a novel failure mode of \textit{class-specific mode collapse}, which occurs when conditional GANs are trained on long-tailed data distribution. Through our analysis we find that the class-specific collapse for each class correlates closely with a sudden increase (explosion) in the spectral norm of its (grouped) conditional BatchNorm  (cBN) parameters. To mitigate the spectral explosion we develop a novel group Spectral Regularizer (gSR), which constrains the spectral norms and alleviates mode collapse. The gSR reduces spectral norms (estimated through power iteration) of grouped parameters and leads to decorrelation of parameters, which enables GAN to effectively improve on long-tailed data distribution without collapse. Our empirical analysis shows that gSR:  a) leads to improved image generation from conditional GANs (by alleviating class-specific collapse), and b) effectively complements exiting regularizers on discriminator to achieve state-of-the-art image generation performance on long-tailed datasets. One of the limitations present in our framework is that it introduces additional hyperparameter $\lambda$ for the regularizer. Developing an hyperparameter free decorrelated parameterization for alleviating class-specific mode collapse is a good direction for future work. We hope that this work leads to further research on improving GANs for real-world long-tailed datasets. \\
\textbf{Acknowledgements}: This work was supported in part by  SERB-STAR Project (Project:STR/2020/000128), Govt. of India and a Google Research Award. Harsh Rangwani is supported by Prime Minister's Research Fellowship (PMRF). We thank Lavish Bansal for help with StyleGAN experiments.

\clearpage

\appendix

\renewcommand \thepart{}
\renewcommand \partname{}

\noindent
\begin{center}
\textbf{\Large Supplementary Material: \\ Improving GANs for Long-Tailed Data through Group Spectral Regularization} 
\end{center}
\renewcommand{\labelitemii}{$\circ$}

This supplementary document is organized as follows:
\begin{itemize}
\setlength{\itemindent}{-0mm}
    \item Section~\ref{sec:supp:notations}: Notations
    \item Section~\ref{sec:supp:metrics}: Additional Metrics
    \item Section~\ref{sec:supp:corr}: Correlations between Spectral Norms and
Class-Specific Mode Collapse
    \item Section~\ref{sec:supp:analysis}: Analysis of Covariance of grouped cBN Parameters
    \item Section~\ref{sec:supp:qual}: Qualitative results
    \item Section~\ref{sec:supp:exp}: Experimental Details
    \begin{itemize}
        \item Datasets (Sec.~\ref{subsec:supp:datasets})
        \item LeCam Regularizer (Sec.~\ref{subsec:supp:lecam})
        \item Spectral Norm Computation Time (Sec.~\ref{subsec:supp:comp})
        \item Sanity Checks (Sec.~\ref{subsec:supp:sanity})
        \item Hyperparameters (Sec.~\ref{subsec:supp:hparams})
        \item Intuition about $n_c$ and $n_g$ (Sec.~\ref{subsec:supp:intuition})
    \end{itemize}
    \item Section~\ref{sec:supp:analysis_gsr}: Analysis of gSR
    \item Section~\ref{sec:supp:sg2}: gSR for StyleGAN2
\end{itemize}{}

\section{Notations}
\label{sec:supp:notations}
\noindent We summarize the notations used in the paper in Table~\ref{tab:supp:notations}.

\begin{table}[!b]{

\vspace{-0.4cm}
\caption{\textbf{Additional metrics on CIFAR-10 dataset.}}
\label{tab:supp:intra-fid}}
\resizebox{\linewidth}{!}
{ 
\begin{tabular}{lccccc|ccccc}
\toprule
Imb. Ratio ($\rho$)                & \multicolumn{5}{c|}{100}                                                                          & \multicolumn{5}{c}{1000}                                                                          \\
                                  & \multicolumn{1}{l}{Intra-class FID} & Precision & Recall & \multicolumn{1}{l}{Density} & Coverage & \multicolumn{1}{l}{Intra-class FID} & Precision & Recall & Density & \multicolumn{1}{l}{Coverage} \\ \cmidrule(l){2-11} 
SNGAN   & 78.36                               & 0.69      & 0.53   & 0.67                        & 0.51     & {121.57}          & 0.60      & \textbf{0.40}    & 0.43    & 0.32                         \\
 \; + gSR (Ours) & \textbf{55.71}                               & \textbf{0.71}      & \textbf{0.56}   & \textbf{0.76}                        & \textbf{0.67}     & \textbf{108.12}                              & \textbf{0.63}      & 0.39   & \textbf{0.53}    & \textbf{0.34}                         \\ \midrule
BigGAN       & 57.82                               & 0.65      & \textbf{0.58}   & 0.63                        & 0.67     & 109.29                              & 0.56     & 0.50   & 0.40    & 0.40                          \\
 \; + gSR (Ours) & \textbf{43.41}                               & \textbf{0.74}      & 0.56   & \textbf{0.93}                        & \textbf{0.80}      & \textbf{98.59}                               & \textbf{0.59}      & \textbf{0.51}   & \textbf{0.49}    & \textbf{0.51}                         \\ \bottomrule
\end{tabular}
}

\vspace{-0.5cm}
\end{table}

\begin{table}[t]
\centering
\caption{\textbf{Notation Table}}
\label{tab:supp:notations}
\resizebox{\linewidth}{!}{%
\begin{tabular}{p{0.1\linewidth} p{0.225
\linewidth} p{0.65\linewidth}}
\toprule
Symbol                & Space                              & Meaning                                                                               \\ \midrule
$K$                     & $\mathbb{N}$                       & Number of Classes                                                                     \\
$y$                     & \{1, 2, ..., $K$\}                     & Class label                                                                           \\
$\mathbf{z}$            & $\mathbb{R}^{256}$                   & Noise vector                                                                          \\
$D$                     &                                    & Discriminator                                                                         \\
$G$                     &                                    & Generator                                                                             \\
$\mathbf{x}$            & $\mathbb{R}^{3 \times H \times W}$ & Image                                                                                 \\
$\mathbf{x^l_y}$        & $\mathbb{R}^d$                     & Feature vector from the Generator's l$^{th}$ cBN's input feature map                    \\
$\mathbf{\mu^l_B}$      & $\mathbb{R}^d$                     & Mean of incoming features to Generator's l$^{th}$ cBN from minibatch $B$                \\
$\mathbf{\sigma^l_B}$ & $\mathbb{R}^d$                     & Std. dev. of incoming features to Generator's l$^{th}$ cBN from minibatch $B$           \\
$\mathbf{\gamma^l_y}$   & $\mathbb{R}^d$                     & Gain parameter for $y^{th}$ class of $l^{th}$ cBN layer of Generator                  \\
$\mathbf{\beta^l_y}$    & $\mathbb{R}^d$                     & Bias parameter for $y^{th}$ class of $l^{th}$ cBN layer of Generator                  \\
$n_g$                   & $\mathbb{R}$                       & Number of groups                                                                      \\
$n_c$                   & $\mathbb{R}$                       & Number of columns                                                                     \\
$\mathbf{\Gamma^l_y}$    & $\mathbb{R}^{n_g \times n_c}$      & $\mathbf{\gamma^l_y}$ after grouping                                                  \\
$\mathbf{B^l_y}$       & $\mathbb{R}^{n_g \times n_c}$      & $\mathbf{\beta^l_y}$ after grouping                                                   \\
$\sigma_{max}$          & $\mathbb{R}^+$                     & Spectral norm                                                                         \\
$n_y$                   & $\mathbb{N}$                       & Number of samples in class $y$                                                        \\
$\rho$                  & $\mathbb{R}$                       & Imbalance ratio: Ratio between the most and the least frequent classes of the dataset \\ \bottomrule
\end{tabular}}
\end{table}

\section{Additional Metrics}
\label{sec:supp:metrics}
In addition to FID and IS reported for experiments in main paper, we also evaluate additional metrics of Precision \cite{kynkaanniemi2019improved}, Recall \cite{kynkaanniemi2019improved}, Density~\cite{yu2020inclusive} and Coverage~\cite{yu2020inclusive} and Intra-FID for CIFAR-10 dataset.
We observe that across all the 4 different imbalance configurations (as in main paper Table {\color{red} 2}) there is significant improvement in all metrics but Recall (which is comparable to baseline in all cases.

\section{Correlations between Spectral Norms and Class-Specific Mode Collapse}
\label{sec:supp:corr}
\begin{figure*}[ht]
     \centering
     \begin{subfigure}[b]{\textwidth}
         \centering
         \includegraphics[width=\textwidth]{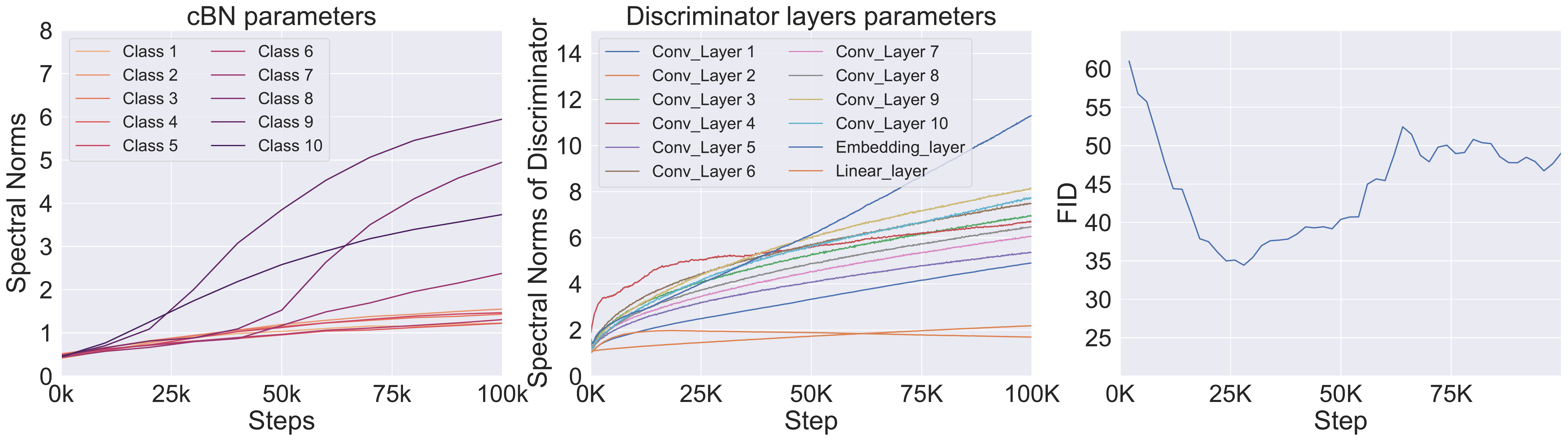}
         \caption{Without gSR}
         \label{fig:supp:wo_gsr}
     \end{subfigure}
     \hfill
     \begin{subfigure}[b]{\textwidth}
         \centering
         \includegraphics[width=\textwidth]{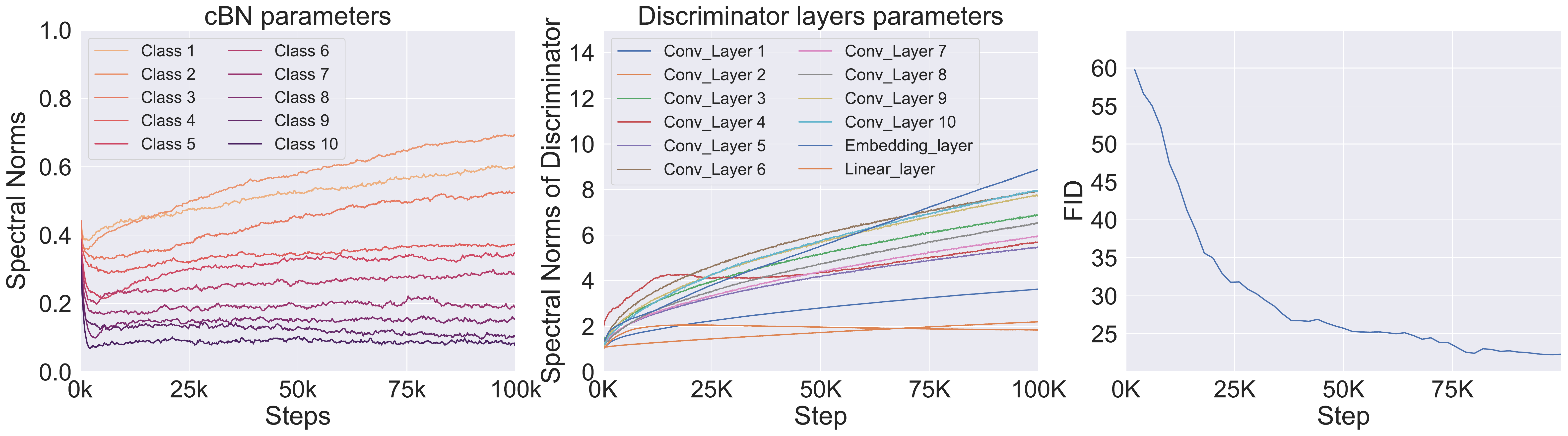}
         \caption{With gSR}
         \label{fig:supp:w_gsr}
     \end{subfigure}
     
    \caption{\textbf{Class-specific mode collapse exhibits unique behaviour with respect to cBN parameters.} Class-specific mode collapse leads to spectral explosion in Generator's cBN parameters' spectral norms (left), which correlates with explosion of FID (right), while having little effect on discriminator's parameters' spectral norms (middle). Class-specific mode collapse is remedied by gSR which keeps the cBN parameters' spectral norms under control.}
        \label{fig:supp:disc_specnorms}
        \vspace{-4mm}
\end{figure*}

In this section, we provide additional details and comparisons to emphasize the differences between class-specific mode collapse and the usual mode collapse (as described in main paper Sec.~\ref{subsec:class-specific-collapse}.
In SNGAN~\cite{miyato2018spectral} and  BigGAN~\cite{brock2018large}, the discriminator's (D) weights' spectral norms tend to explode as the mode collapse occurs for balanced data. To determine if this also occurs in long-tailed case we train a SNGAN on CIFAR-10 ($\rho$ = 100) (with and without gSR) and plot the spectral norm of weights of discriminator layers. We find that spectral explosion for discriminator weights is not observed in the class-specific mode collapse (without gSR case), as we report in Fig.~\ref{fig:supp:disc_specnorms}. Discriminator's layers' spectral norms do not show significant change before and after applying gSR . On the other hand, before applying gSR the spectral norms of class-specific parameters of cBN explode (at step 25k and 50k). At the same stage FID suddenly increases, whereas there is no anomaly in Discriminator's spectral norms'.  Thus, the class-specific mode collapse behaviour is different as compared to that of the mode collapse previously reported in the literature \cite{miyato2018cgans, brock2018large}, and cannot be detected through discriminator spectral norms. Hence, it's detection requires the analysis of spectral norms of grouped parameters in cBN  which we propose in this paper. 
 
 The above spectral explosion of the generator's cBN motivates us to formulate gSR (Sec.~\ref{sec:regularizer}). We find (Fig. \ref{fig:supp:disc_specnorms}) that after applying gSR there is no spectral collapse and training is stabilized (decreasing FID).

\section{Analysis of Covariance of grouped cBN Parameters}
\label{sec:supp:analysis}
\begin{figure*}[!ht]
     \centering
     \begin{subfigure}[b]{\textwidth}
         \centering
         \includegraphics[width=\textwidth, height=140pt]{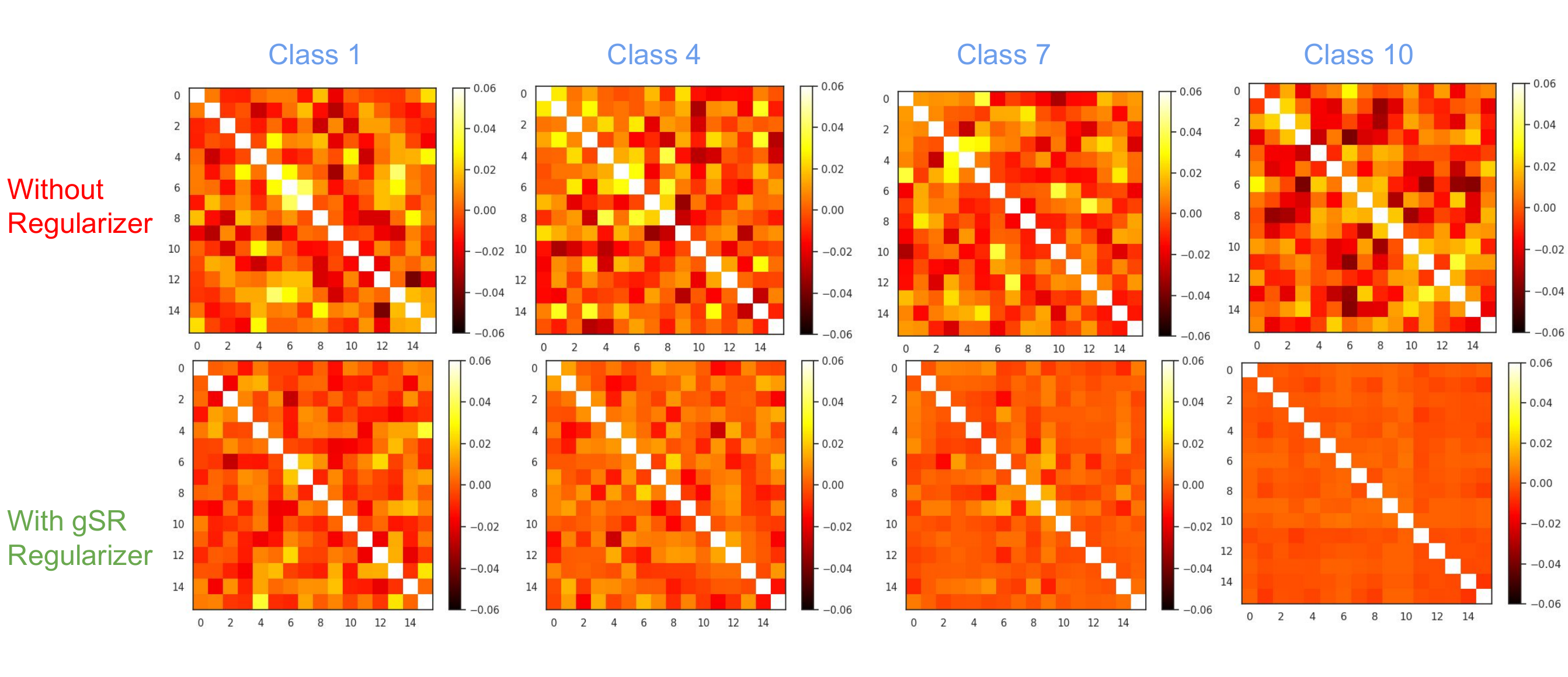}\\[-3ex]
         \caption{Covariance matrices of $\mathbf{\Gamma_y^l}$ for \textbf{(l = 1)} for SNGAN baseline.}
         \label{fig:supp:y equals x}
     \end{subfigure}
     \hfill
     \begin{subfigure}[b]{\textwidth}
         \centering
         \includegraphics[width=\textwidth, height=140pt]{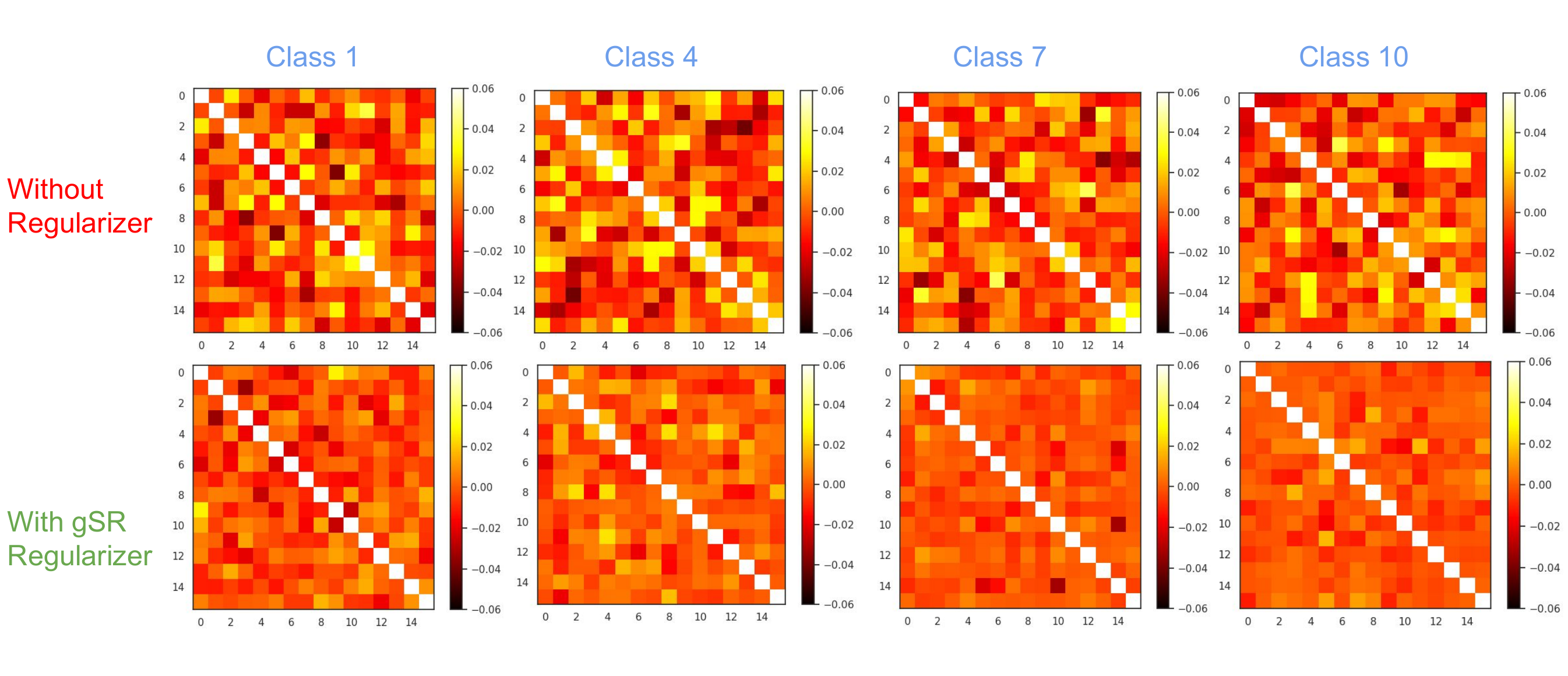}\\[-3ex]
         \caption{Covariance matrices of $\mathbf{\Gamma_y^l}$ for \textbf{(l = 3)} for SNGAN baseline.}
         \label{fig:supp:three sin x}
     \end{subfigure}
     \hfill
     \begin{subfigure}[b]{\textwidth}
         \centering
         \includegraphics[width=\textwidth, height=140pt]{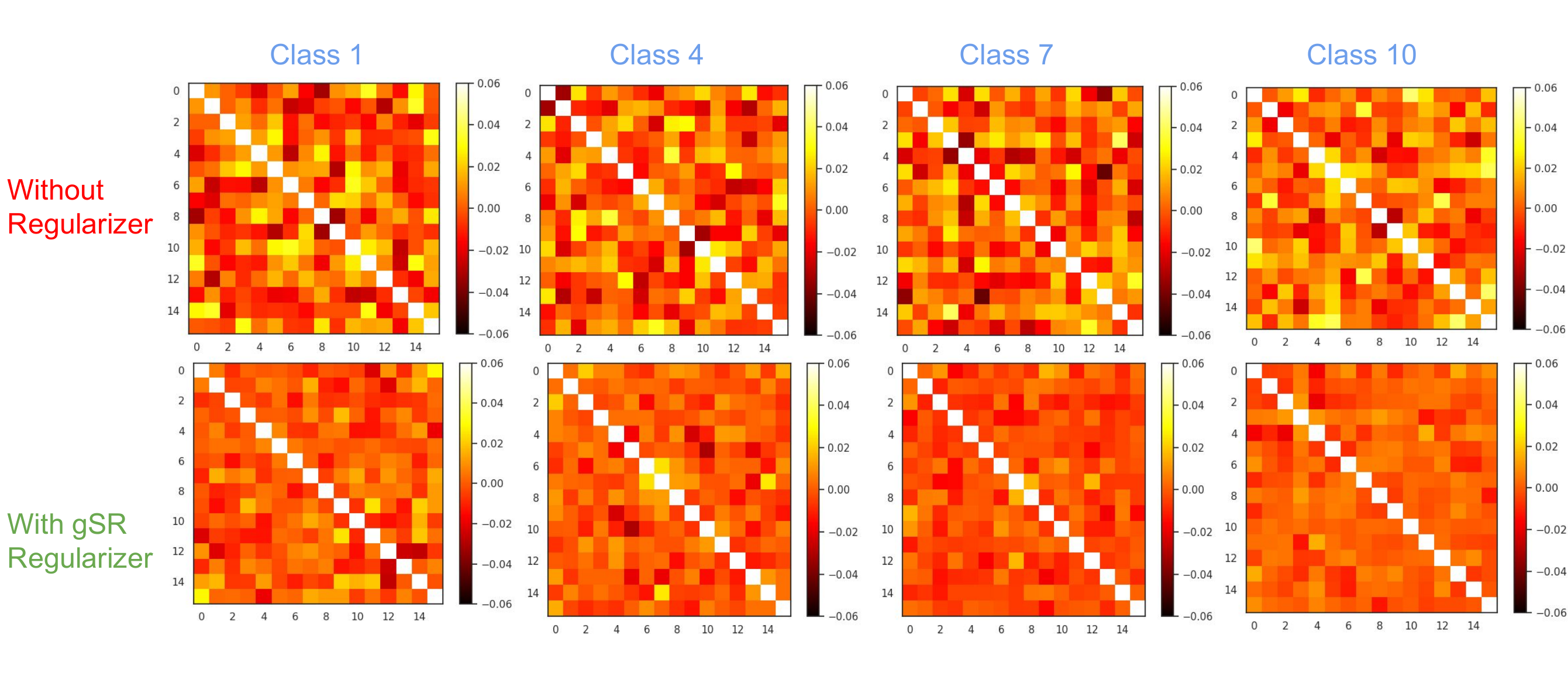}\\[-3ex]
         \caption{Covariance matrices of $\mathbf{\Gamma_y^l}$ for \textbf{(l = 5)} for SNGAN baseline.}
         \label{fig:supp:five over x}
     \end{subfigure}
    \caption{\textbf{Covariance matrices of $\mathbf{\Gamma_y^l}$ for SNGAN baseline on CIFAR-10 ($\rho = 100$).}}
    \label{fig:supp:three graphs}
\end{figure*}
\vspace{-3mm}
For analyzing the decorrelation effect of gSR (explained in Sec.~\ref{sec:regularizer}), we train a SNGAN on CIFAR-10 ($\rho$=100) with gSR. 
We then visualize the covariance matrices of $\mathbf{\Gamma^l_y}$ (grouped $\mathbf{\gamma^l_y}$) across cBN at different layers $\mathbf{l}$ in the generator. gSR leads to suppression of covariance between off-diagonal features of $\mathbf{\Gamma^l_y}$ belonging to the tail classes, implying decorrelation of parameters (Sec.~\ref{sec:regularizer}). As we go from initial to final cBN layers of the Generator, we see that this suppression is reduced in the case when gSR is applied. This leads to increased similarity between the covariance matrices of the head class and tail class. This effect can be attributed to the features learnt at the respective layers. The initial layers (in G) are responsible for more abstract and class-specific features, whereas the final layers produce features while are more fine-grained and generic across different classes. This is in contrast to what is observed for a classifier, as the generator is an inverted architecture in comparison to a classifier.

\section{Qualitative Results}
\label{sec:supp:qual}
\begin{figure*}[t]    
    \centering
    \includegraphics[width=\textwidth]{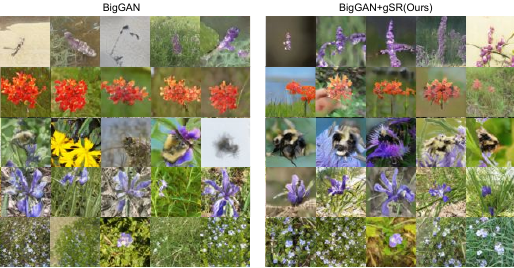}
    \caption{\textbf{Qualitative comparison of BigGAN variants on Tail classes from iNaturalist 2019 dataset ($\rho$=100) (64 $\times$ 64).} Each row represents images from a distinct class. }
    \label{fig:supp:inat_64}
    \vspace{-3mm}
\end{figure*}

\begin{figure*}[!t]
    \centering
    \includegraphics[width=\linewidth]{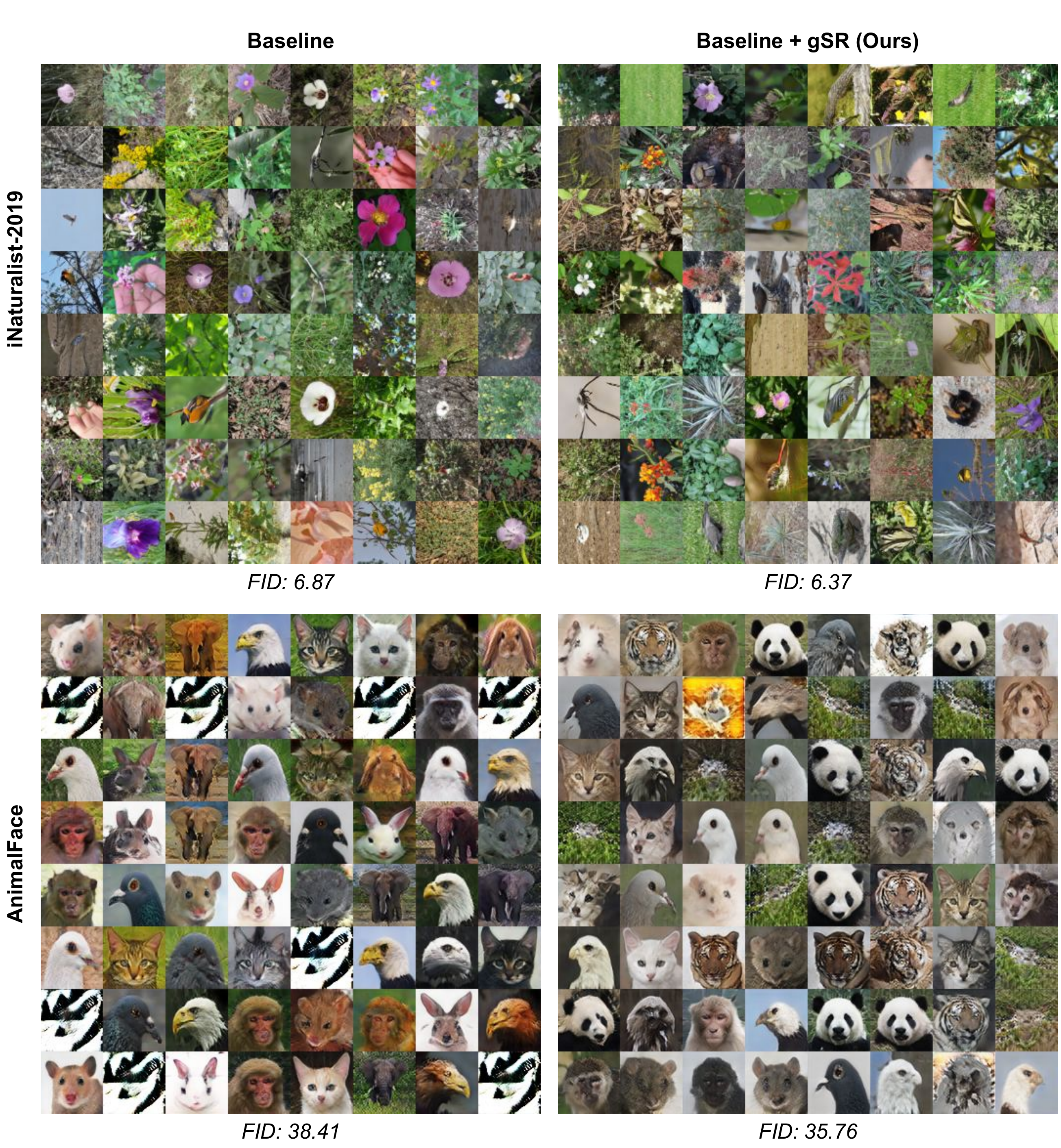}
    \caption{\textbf{Qualitative Results.} The baseline is composed of BigGAN~\cite{brock2018large}+LeCam~\cite{tseng2021regularizing}+DiffAug~\cite{zhao2020differentiable}. gSR improves the quality and diversity of the images generated by baseline over challenging iNaturalist-19 and AnimalFace datasets.}
    \label{fig:supp:qual_results_supp}
\end{figure*}

We show generated images on iNaturalist-2019 and AnimalFace in Fig.~\ref{fig:supp:qual_results_supp} and Fig. \ref{fig:supp:inat_64}. These are naturally occurring challenging data distributions for training a GAN. Sample diversity as well as quality is improved after applying our gSR regularizer. We also provide a video showing class specific collapse for BigGAN for CIFAR-10 in \texttt{gSR.mp4}.

\section{Experimental Details}
\label{sec:supp:exp}
In this section, we elaborate on the technical and implementation details provided in Sec.~\ref{sec:expts} of the main paper.
\subsection{Datasets}
\label{subsec:supp:datasets}
We describe the datasets used in our work below:

\vspace{1mm}\noindent\textbf{CIFAR-10}: We use CIFAR-10 \cite{krizhevsky2009learning} dataset which comprises of 32 $\times$ 32 images. The dataset is split into $50$k training images and $10$k test images. We use the training images for GAN training and the $10$k test set for calculation of FID. 

\vspace{1mm}\noindent\textbf{LSUN}:
We use a 250k subset of LSUN~\cite{journals/corr/YuZSSX15} dataset as followed by \cite{rangwani2021class, santurkar2018classification}, which is split across the classes of bedroom, conference room, dining room, kitchen and living room classes. We use a balanced subset of 10k images balanced across classes for FID calculation.

\vspace{1mm}\noindent\textbf{iNaturalist-2019}:
The iNaturalist-2019~\cite{inat19} is a long-tailed dataset composed of 268,243 images present across 1010 classes in the training set. The validation set contains 3030 images balanced across classes, used for FID calculation.

\vspace{1mm}\noindent\textbf{AnimalFace~\cite{si2011learning}}: The AnimalFace dataset contains 2,200 RGB images across 20 different categories with images containing animal faces. We use the training set for calculation of FID as there is no seperate validation set provided for baselines. Our results on this dataset show that our regularizer can also help in preventing collapse in extremely low data (i.e. few shot) scenario's as well.

\subsection{LeCam Regularizer}
\label{subsec:supp:lecam}
We use LeCam regularizer~\cite{tseng2021regularizing} for all our experiments.
{\small
\begin{align}
R_{LC} = \mathop{\mathbb{E}}_{\mathbf{x}\sim\mathcal{T}}[\Vert D(\mathbf{x})-\alpha_F\Vert^2] + \mathop{\mathbb{E}}_{\mathbf{z}\sim p_{\mathbf{z}}}[\Vert D(G(\mathbf{x}))-\alpha_R\Vert^2]
\label{eq:lc_reg}
\vspace{-1mm}
\end{align}
}
\noindent \vspace{1mm} LeCam regularizer computes exponential moving average of discriminator outputs for real and generated images. The difference between discriminator outputs for real and generated images is taken against the moving averages of discriminator outputs of generated images ($\alpha_F$) and real images ($\alpha_R$) respectively. This does not allow the discriminator to output predictions with very high confidence, thereby preventing overfitting by keeping the predictions in a particular range. We use the $\lambda_{LC}$ value of 0.1, 0.3 and 0.01 as suggested by the authors \cite{tseng2021regularizing} ,which is specified in Table~\ref{tab:supp:hyper}. The term $\lambda_{LC}R_{LC}$ is then added to discriminator loss for regularization.

\subsection{Spectral Norm Computation Time}
\label{subsec:supp:comp}
Since our regularizer involves estimating largest singular value for $\mathbf{{\Gamma}_y^l}$, this can be done through either power iteration or SVD. We use power iterations method to calculate the singular values of $\mathbf{\Gamma^l_y}$ and $\mathbf{B^l_y}$. We use 4 power iterations for estimating the largest singular value. For perfect decorrelation, other techniques like Group Whitening \cite{huang2021group} can also be used, but they involve full SVD computation. We provide a comparison of time for 100 generator steps of training for baseline, baseline (w/ power iteration (piter)) and baseline (with full SVD) computation for iNaturalist 2019 dataset in table below. All the runs were done on NVIDIA RTX 3090 GPU on the same machine.  
\vspace{2mm}

\begin{table}[h]
\centering
\begin{tabular}{l|c}
\toprule
                  & Time (in secs) \\ \midrule
BigGAN            & 68             \\ 
BigGAN (w/ piter) & 77             \\ 
BigGAN (w/ SVD)   & 1126         \\ \bottomrule
\end{tabular}
\caption{Comparison of time taken for 100 updates of generator(G) on iNaturalist-2019 dataset. }
\label{tab:supp:time_comp}
\end{table}

 As for each class separate SVD computation is performed we find that the SVD computation becomes very expensive (Table~\ref{tab:supp:time_comp}) for large datasets like iNaturalist-2019. Whereas as the power iteration can be done in parallel there is not much computation overhead with addition of each class. Hence, techniques like Group Whitening \cite{huang2021group} which use SVD are not a viable baseline for our case. It can be observed that despite having large number of classes in iNaturalist there is only addition of 9 sec, which shows the scalability and viability of proposed gSR. We provide a PyTorch implementation of cBN, detailing the process of spectral norm calculation as part of the supplemental material.

\subsection{Sanity Checks}
\label{subsec:supp:sanity}

We build our experiments over the PyTorch-StudioGAN framework, which provides a simple framework over standard GAN architectures and setups. Since we are not using the official code for the LeCam Regularizer baseline \cite{tseng2021regularizing}, we first reproduce the BigGAN (+ LeCam + DiffAug) results on CIFAR-10 to ensure that our codebase is on par with the official codebase of the LeCam GAN. Our code obtains an FID of 7.59$_{\pm 0.04}$ vs. 8.31$_{\pm 0.03}$ reported in same setting by \textit{Tseng}~\etal~\cite{tseng2021regularizing}, which verifies the authenticity of our experiments. Hence, we compare our results to a stronger baseline which is due to improved implementation of BigGAN in the framework.

\begin{table}[!t]
\centering
\caption{Hyperparameter setups for all the reported experiments. $\alpha_D$, and $\alpha_G$ denote the learning rates for Discriminator and Generator respectively.}
\label{tab:supp:hyper}
\resizebox{\linewidth}{!}{
\begin{tabular}{@{}lllllll}
\toprule
\multirow{2}{*}{Setting} & Adam & \multirow{2}{*}{n$_{dis}$} & \multirow{2}{*}{$\lambda_{LC}$} & \multirow{2}{*}{G$_{EMA}$} & EMA & Total \\ 
& ($\alpha_D$, $\alpha_G$, $\beta_1$, $\beta_2$) & & & & Start & Iterations \\ \midrule
A      & 2e-4, 2e-4, 0.5, 0.9                                & 5         & 0.3            & False     &           & 120k             \\
B      & 2e-4, 2e-4, 0.5, 0.999                              & 5         & 0.1            & True      & 1k        & 120k             \\
C      & 2e-4, 2e-4, 0.5, 0.9                                & 5         & 0.3           & True      & 1k        & 200k             \\
D       & 2e-4, 2e-4, 0.0, 0.999                                & 2         & 0.01            & True     & 20k           & 120k             \\
E       & 2e-4, 2e-4, 0.5, 0.999                              & 5         & 0.01           & True      & 1k        & 120k             \\ 
F       & 4e-4, 1e-4, 0.5, 0.9                                & 5         & 0.5             & True     & 1k           & 120k             \\

\bottomrule
\end{tabular}}

\bigskip 

\resizebox{\linewidth}{!}{
\begin{tabular}{l|c|c|c|c}
    \toprule
    & CIFAR-10 & LSUN & iNaturalist-19 & AnimalFace\\ \midrule
    LSGAN~\cite{mao2017least} & \multicolumn{2}{c|}{\multirow{3}{*}{A}} & \multicolumn{2}{c}{\multirow{3}{*}{---}} \\
    SNGAN~\cite{miyato2018spectral} & \multicolumn{2}{c|}{} & \multicolumn{2}{c}{}\\
    \; + gSR (Ours) & \multicolumn{2}{c|}{} & \multicolumn{2}{c}{} \\ \hline
    BigGAN~\cite{brock2018large} & \multirow{2}{*}{B}  & \multirow{2}{*}{C|F} & \multirow{2}{*}{D}  & \multirow{2}{*}{E}\\
     \; + gSR (Ours) &  &  & &\\ \bottomrule
    \end{tabular}}

\end{table}

\subsection{Hyperparameters}
\label{subsec:supp:hparams}
We provide the details of the hyperparameters used in the experiments in Table~\ref{tab:main_results} and ~\ref{tab:iNaturalist} of the main paper in Table~\ref{tab:supp:hyper}. For CBGAN~\cite{rangwani2021class} based experiments we follow the same setup as reported in the paper (except using a ResNet~\cite{gulrajani2017improved} architecture for fairness in experiments). For BigGAN on LSUN dataset we use configuration C for the imbalance factor ($\rho$ = 100) and F for imbalance factor ($\rho$ = 1000). In our tuning experiments we explored the configurations in Table \ref{tab:supp:hyper} and use the configuration which produces best FID for baseline. Then we add gSR regularizer to obtain our results. 

\noindent \vspace{1mm} \textbf{High-Resolution Experiments:} For the high resolution ($128 \times 128$) image synthesis on LSUN we find that we only require very small change in hyperparameters for obtaining results. For SNGAN, we use configuration A in Table~\ref{tab:supp:hyper} with EMA starting at 1k along with $\lambda_{LC} = 0.5$. For the BigGAN we use the same configuration as in the Table \ref{tab:supp:hyper}. We find that for higher resolutions a larger $\lambda_{LC}$ helps the purpose.

\subsection{Intuition about $n_c$ and $n_g$}
\label{subsec:supp:intuition}
As we group the parameters $\gamma_{\mathbf{y}}^{\mathbf{l}}$ (Eq. {\color{red} 3} in main paper) to a matrix $\mathbf{\Gamma_{\mathbf{y}}^{\mathbf{l}}}$ of $n_c \times n_g$. The matrix can be decomposed into $\min(n_c,n_g)$ (matrix rank) number of independent and diverse components through SVD. As the scope of attaining maximal orthogonal and diverse components (matrix rank) is when $n_c \approx n_g$, it helps gSR to ensure maximal diversity and performance (as seen in main paper Table {\color{red} 5}). In case of gSR we find that almost all eigen values of $\mathbf{\Gamma_{\mathbf{y}}^{\mathbf{l}}}$ have a similar value, which demonstrates orthogonality and diversity. \\

\section{Analysis of gSR}
\label{sec:supp:analysis_gsr}

\vspace{1mm}\noindent\textbf{How much should be gSR's strength ($\lambda_{gSR}$)? }
\setlength{\intextsep}{0pt}%
\begin{wrapfigure}{r}{0.5\textwidth}
    \includegraphics[width=0.5\textwidth]{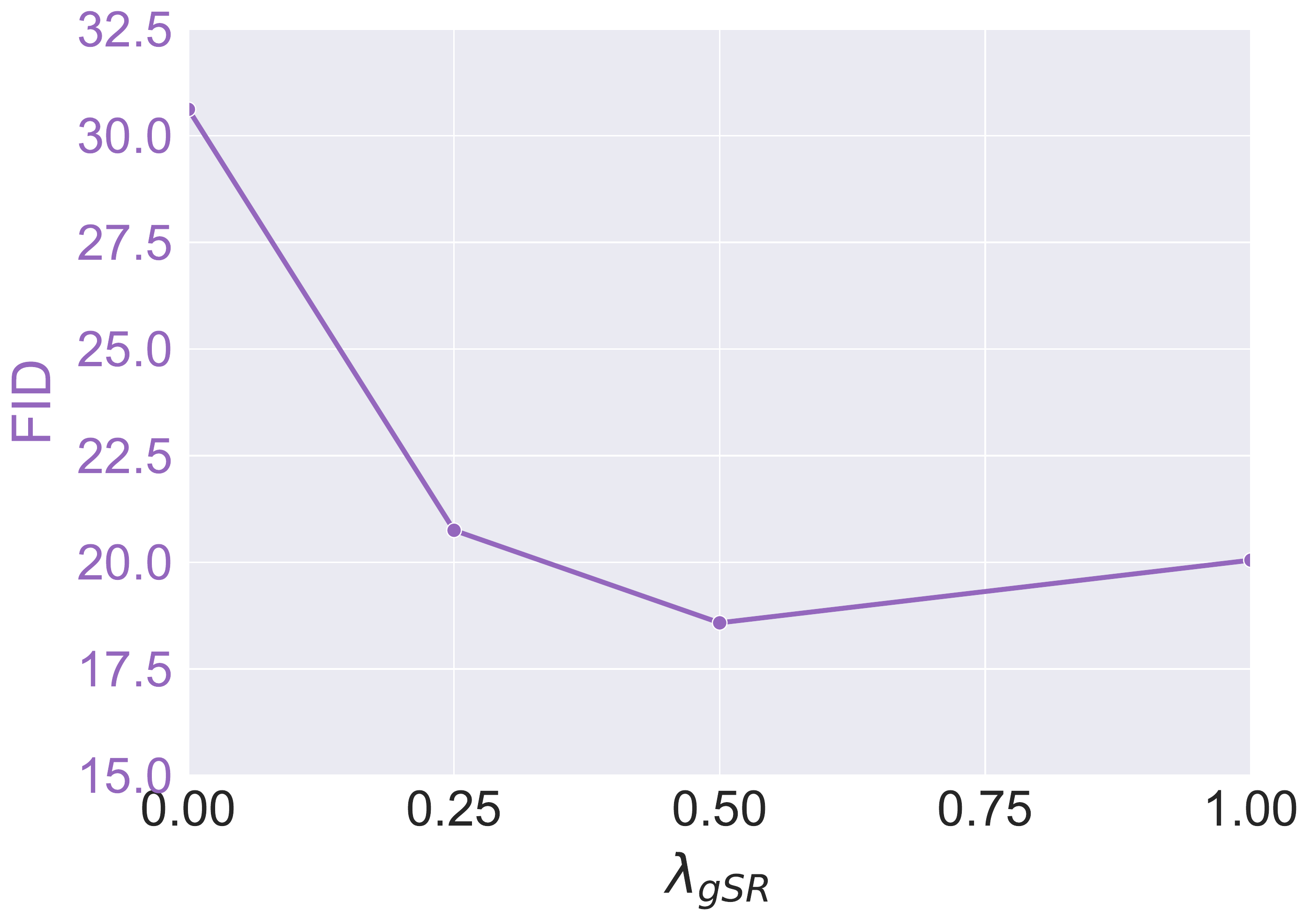}
    \caption{\textbf{Sensitivity to $\lambda_{gSR}$.}  On CIFAR-10, the FID marginally changes with $\lambda_{gSR}$ (0.25 to 1).}
    \label{fig:supp:fid_lambda}
\vspace{-8mm}
\end{wrapfigure} 

\noindent \vspace{1mm} We experiment with different values of $\lambda_{gSR}$ for gSR in SNGAN as shown in Fig.~\ref{fig:supp:fid_lambda}. $\lambda_{gSR}$ value of 0.5 attains best FID scores, hence we use it for all our experiments. 
 The value of FID changes marginally when $\lambda_{gSR}$ goes from 0.25 to 1 which highlights its robustness (\ie less sensitivity).

\section{gSR for StyleGAN2}
\label{sec:supp:sg2}
\begin{figure}[t]
    \centering
  \includegraphics[width=\linewidth]{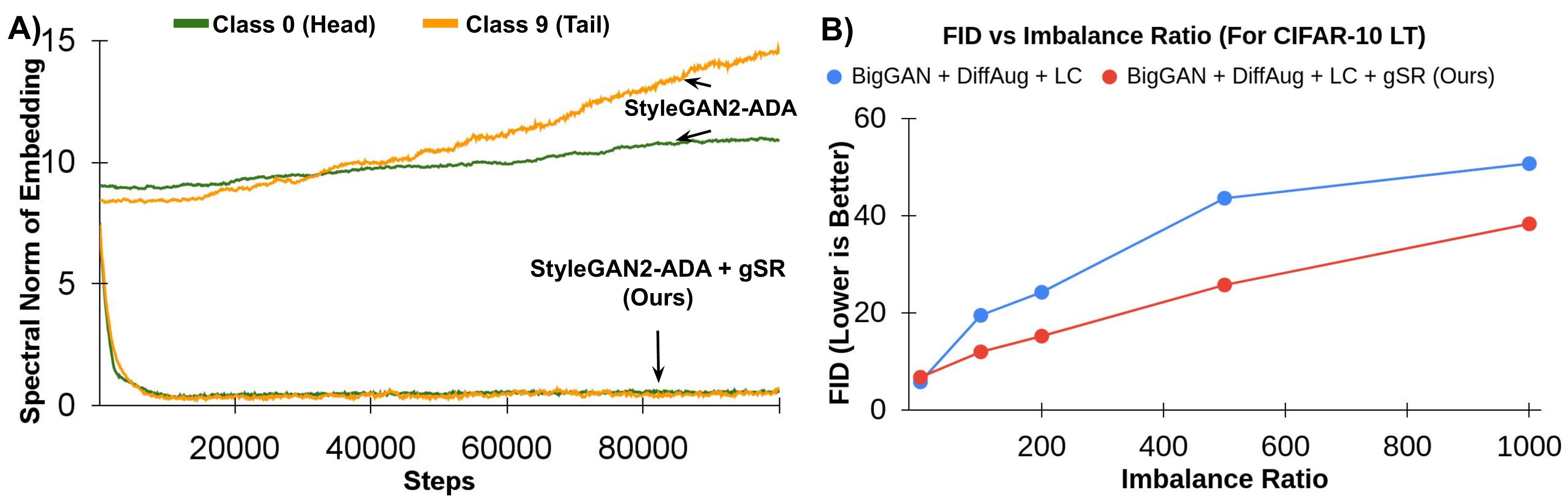}
    \caption{\small\textbf{A)} Spectral Norm of class embeddings used in conditional StyleGAN2-ADA.\textbf{B)} Mean FID vs Imbalance ratio.}
    \label{fig:supp:ablation}
\end{figure}
 \begin{table}[t]
    \centering
    \caption{Quantitative comparison of  gSR over StyleGAN2-ADA baseline.}
    \label{tab:quant_sg2}
    \begin{tabular}{lcccc} \hline
     & \multicolumn{2}{c}{CIFAR10-LT ($\rho = 100$)} & \multicolumn{2}{c}{LSUN-LT ($\rho = 100$)} \\ \hline
    & FID-10k ~\cite{zhao2020differentiable}  ($\downarrow$)&      IS($\uparrow$)& FID-10k ~\cite{zhao2020differentiable} ($\downarrow$)&      IS($\uparrow$)  \\\hline
  StyleGAN2-ADA & 71.09$_{\pm0.12}$ & 5.66$_{\pm0.03}$ & 55.04$_{\pm0.07}$ & 3.92$_{\pm0.02}$ \\
  +gSR(Ours) & \textbf{22.76}$_{\pm0.17}$ & \textbf{7.55}$_{\pm0.01}$ & \textbf{27.85}$_{\pm0.06}$ & \textbf{4.32}$_{\pm0.01}$ \\ \hline
  \end{tabular}
  \label{tab:stylegan2}
\end{table}
\begin{figure}[h]
    \centering
  \includegraphics[width=0.8\linewidth]{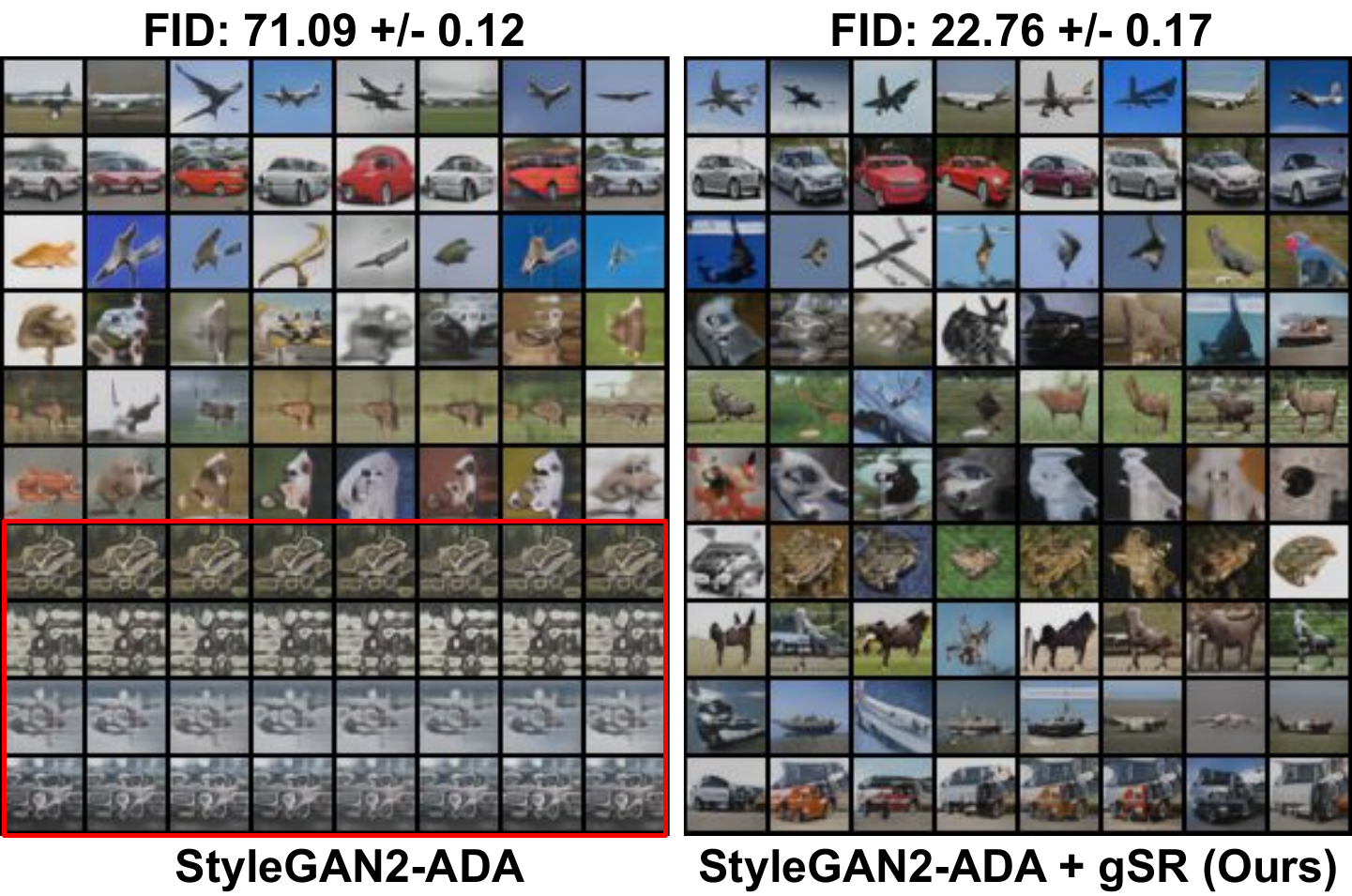}
    \caption{\small Qualitative comparison of baseline and baseline + gSR on ImageNet-LT (\emph{left}) and CIFAR10-LT (\emph{right}).}
    \label{fig:supp:qual_results}
\end{figure}
 We train and analyze the spectral norm of class-conditional embeddings in StyleGAN2-ADA implementation available ~\cite{kang2020contrastive} on long-tailed datasets (CIFAR10 and LSUN), to find that it also suffers from spectral collapse of tail class embedding parameters (Fig.\ \ref{fig:supp:ablation}) as BigGAN and SNGAN. We then implement gSR on StyleGAN2 generator by grouping 512 dimensional class conditional embeddings to 16x32 and calculating their spectral norm  which is added to loss (Eq.\ \ref{eq:reg_loss}) as $R_{gSR}$. We find that gSR is able to effectively prevent the mode collapse (Fig.~\ref{fig:supp:qual_results}) and also results in significant improvement in FID (Table \ref{tab:stylegan2}) in comparison to StyleGAN2-ADA baseline.

\clearpage

\bibliographystyle{splncs04}
\bibliography{egbib}
\end{document}